\pdfoutput=1

\documentclass[11pt]{article}

\usepackage[]{acl}

\usepackage{times}
\usepackage{latexsym}

\usepackage[T1]{fontenc}

\usepackage{CJKutf8}
\usepackage[utf8]{inputenc}

\usepackage{microtype}

\usepackage{soul}
\usepackage{subfigure}
\usepackage{bm}
\usepackage{amsmath}
\usepackage{amssymb}
\usepackage{tablefootnote}
\usepackage{multirow}
\usepackage{booktabs}
\usepackage{arydshln}
\usepackage{tikz}
\usetikzlibrary[arrows.meta, decorations.pathmorphing, backgrounds, positioning, fit, petri, graphs, math, external, calc, shapes, shapes.geometric, decorations.text, shadings]
\usepackage{tikz-dependency}
\usepackage{xcolor}
\usepackage{microtype}
\usepackage{mathtools}
\usepackage{adjustbox}
\usepackage{enumitem}

\setlist{nosep}

\definecolor{brickred}{HTML}{b92622}
\definecolor{midnightblue}{HTML}{005c7f}
\definecolor{salmon}{HTML}{f1958d}
\definecolor{burntorange}{HTML}{f19249}
\definecolor{junglegreen}{HTML}{4dae9d}
\definecolor{forestgreen}{HTML}{499c5e}
\definecolor{pinegreen}{HTML}{3d8a75}
\definecolor{seagreen}{HTML}{6bc1a2}
\definecolor{limegreen}{HTML}{97c65a}
\newcommand{\red}[1]{\textcolor{brickred}{#1}}
\newcommand{\blue}[1]{\textcolor{midnightblue}{#1}}

\pgfdeclarepatternformonly{south west lines}{\pgfqpoint{-0pt}{-0pt}}{\pgfqpoint{3pt}{3pt}}{\pgfqpoint{3pt}{3pt}}{
        \pgfsetlinewidth{0.4pt}
        \pgfpathmoveto{\pgfqpoint{0pt}{0pt}}
        \pgfpathlineto{\pgfqpoint{3pt}{3pt}}
        \pgfpathmoveto{\pgfqpoint{2.8pt}{-.2pt}}
        \pgfpathlineto{\pgfqpoint{3.2pt}{.2pt}}
        \pgfpathmoveto{\pgfqpoint{-.2pt}{2.8pt}}
        \pgfpathlineto{\pgfqpoint{.2pt}{3.2pt}}
        \pgfusepath{stroke}}

\title{Fast and Accurate End-to-End Span-based Semantic Role Labeling as Word-based Graph Parsing}

\author{
    \textbf{Shilin Zhou},
    \textbf{Qingrong Xia},
    \textbf{Zhenghua Li}\Thanks{$~$ Corresponding author},
    \textbf{Yu Zhang},
    \textbf{Yu Hong},
    \textbf{Min Zhang} \\
    Institute of Artificial Intelligence, School of Computer Science and Technology, \\
    Soochow University, Suzhou, China \\
    \texttt{\{slzhou.cs,yzhang.cs\}@outlook.com; kirosummer.nlp@gmail.com} \\
    \texttt{\{zhli13,hongy,minzhang\}@suda.edu.cn}
}

\begin{document}
\begin{CJK}{UTF8}{gkai}
\maketitle
\begin{abstract}

This paper proposes to cast end-to-end span-based SRL as a  word-based graph parsing task. 
The major challenge is how to represent spans at the word level. 
Borrowing ideas from research on Chinese word  segmentation and named entity recognition, 
we propose and compare four different schemata of graph representation, i.e., \emph{BES}, \emph{BE}, \emph{BIES}, and \emph{BII}, 
among which we find that the \emph{BES} schema performs the best. 
We further gain interesting insights through detailed analysis.  
Moreover, we propose a simple constrained Viterbi procedure to ensure the legality of the output  graph according to the constraints of the SRL structure. 
We conduct experiments on two widely used benchmark datasets, i.e., CoNLL05 and CoNLL12. 
Results show that our word-based graph parsing approach achieves consistently better performance than previous results, under all settings of end-to-end and predicate-given, without and with pre-trained language models (PLMs). 
More importantly, our model can parse  669/252 sentences per second, without and with PLMs respectively.  
\end{abstract}

\section{Introduction}

As a fundamental natural language processing (NLP) task, semantic role labeling (SRL) uses predicate-argument structure to represent the shallow semantic meaning of sentences.  
SRL structure is shown to be helpful for many downstream NLP tasks, such as machine translation \citep{liu-2010-semantic, marcheggiani-2018-exploiting} and question answering \citep{wang-2015-machine}. 

\begin{figure}[tb]
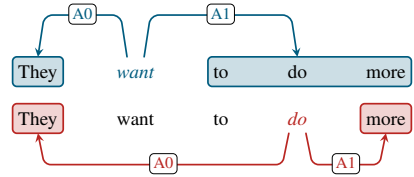

    \centering
    \scalebox{0.8}{
    \begin{minipage}[t]{1\linewidth}
            \centering
            \begin{dependency}[]
                \begin{deptext}[row sep=0.3cm, column sep=0.8cm, font=\small]
                            They \& \blue{{\emph{want}}} \& to \& do \& more \& . \\
                            They \& want \& to \& \red{\emph{do}} \& more \& . \\
                \end{deptext}
                \depedge[edge style={midnightblue,thick}, edge height=4ex]{2}{1}{\blue{A0}}
                \depedge[edge style={midnightblue,thick}, edge height=4ex]{2}{4}{\blue{A1}}
                \depedge[edge below,edge style={brickred,thick},edge height=3ex]{4}{1}{\red{A0}}
                \depedge[edge below,edge style={brickred,thick},edge height=3ex]{4}{5}{\red{A1}}
               \wordgroup[color=midnightblue,thick,fill=midnightblue!20]{1}{1}{1}{prd1-a0}
              \wordgroup[color=midnightblue,thick,fill=midnightblue!20]{1}{3}{5}{prd1-a1}
               
              \wordgroup[color=brickred,thick,fill=brickred!20]{2}{1}{1}{prd2-a0}
              \wordgroup[color=brickred,thick,fill=brickred!20]{2}{5}{5}{prd2-a1}
            \end{dependency}
        \end{minipage}}%
    \caption{An example of span-based SRL, where ``\emph{want}'' and ``\emph{do}'' are two predicates. 
    } 
    \label{fig:raw-srl}
\end{figure}

There exist two forms of concrete SRL formalism in the community, i.e., word-based (also known as dependency-based SRL)
and span-based, depending on whether an argument consists of a single word or a word span. 
Compared  with word-based SRL, span-based SRL is more complex due to difficulties in determining argument boundaries. 
Figure \ref{fig:raw-srl} shows the span-based SRL structure for two predicates.  
Semantic roles of arguments are distinguished with edge labels, such as ``\texttt{A0}'' (agent) and ``\texttt{A1}'' (patient). 
This work focuses on the end-to-end  span-based SRL task, and proposes a unified model to simultaneously recognize predicates and arguments in the input sentence.  

In recent years, span-based SRL has achieved substantial performance boost due to the tremendous progress made by deep neural network models, especially by pre-trained  language models (PLMs).
Currently, there are mainly two mainstream  approaches, i.e., 
 BIO-based \cite{zhou-2015-end} sequence labeling and span-based graph parsing  \cite{he-2018-jointly}.

The BIO-based sequence labeling approach first identifies the predicates and then finds arguments for each predicate independently by labeling 
every word with BIO tags, like ``\texttt{B-A0}'' or ``\texttt{I-A0}''. %
Its major weakness 
is that 
a sentence has to be encoded and decoded for multiple times, each time for one predicate \citep{zhou-2015-end, shi-2019-simple}, thus proportionally reducing the training and inference efficiency.\footnote{Some BIO-based approaches, for example \citet{strubell-2018-linguistically}, only encode the input sentence once  without using predicate indicators, but this leads to inferior performance. } 
\citet{zhou-2015-end} concatenate an indicator embedding to each input token, where the focused predicate corresponds to 1, and others to 0.  
\citet{shi-2019-simple} append the focused predicate word to the end of the sentence before getting into BERT \cite{devlin-2018-bert}. 

The span-based graph parsing approach   directly considers all word spans as candidate argument nodes and links them to predicate nodes \citep{he-2018-jointly, Li-2019-dependency}. 
However, this approach also suffers from a severe inefficiency problem, since there are $O(n)$ candidate predicates and $O(n^2)$ candidate arguments, leading to a big search space of 
$O(n^3)$. 
Previous works usually employ heuristic pruning techniques to improve efficiency.

Inspired by recent works on semantic dependency graph parsing (SDGP)
\cite{oepen-2014-semeval,dozat-2018-simpler,tukewei-2019-2osdp}, this work for the first time proposes a word-based graph parsing approach for end-to-end span-based SRL. 
End-to-end means that all predicates and arguments in a sentence are inferred simultaneously and by a single  model.  
The key challenge is how to represent span-based arguments in word-based graphs in which nodes correspond to single words. 
Once this is solved, we can build our parser on the shoulder of existing word-based graph parsing models. This work employs the second-order %
model of \citet{tukewei-2019-2osdp}.  %
In summary, our work makes the following contributions:
\begin{itemize}[leftmargin=*]
    \item
We propose a new word-based graph parsing approach for end-to-end span-based SRL.  Via a straightforward  simplification, our approach can be applied to the \emph{predicate-given} setting.  
    \item Borrowing ideas from research on Chinese word segmentation (CWS) and named entity recognition (NER), we propose and investigate several graph schemata. We find the \emph{BES} schema is steadily superior to others and obtain interesting insights via detailed analysis. 
    \item 
    Inevitably, graph parsing models may output illegal graph that cannot be properly transformed into SRL structure. %
    To deal with this, we propose a simple constrained Viterbi procedure for post-processing illegal graphs. 
    \item
    We conduct experiments on the CoNLL05 and CoNLL12 benchmark datasets.  
    Our proposed approach achieves consistently better performance than previous results, under all settings of end-to-end and predicate-given, with and without PLMs. 
    More importantly, our parser is  much more faster than previous parsers and can analyze 669/252 sentences per second, without and with PLMs. 
\end{itemize}

We release our code, configuration files, and models at \url{https://github.com/zsLin177/SRL-as-GP}.

\section{Related Works} \label{related work}

\paragraph{Span-based SRL.}
As two mainstream neural models, the BIO-based and span-based graph parsing approaches handle SRL in different ways.

\emph{The BIO-based approach} usually predicts predicates first and then recognizes arguments for each predicate via sequence labeling. 
For each predicate,  \citet{zhou-2015-end} indicates the position of the predicate via indicator embedding, and then encode the sentence using multi-layer BiLSTMs, and finally apply a CRF layer to find the best label sequence. 
\citet{shi-2019-simple} append the focused predicate word to the original sentence, and then feed the sentence into BERT, and then apply BiLSTM for further encoding.

\emph{The span-based graph parsing approach} is proposed by \citet{he-2018-jointly}. The idea is directly predicting relations between candidate predicates (single words) and arguments (word spans) in a graph. 
\citet{Li-2019-dependency} apply the approach to the word-based SRL task.

Besides the two mainstream approaches, researchers have explored other interesting directions. 
\citet{zhang-comparing-spanextra} %
propose a two-step span recognition approach, i.e., first identifying a head word and then extending the word into a span. %
\citet{generate-ijcai2021-0521} cast the SRL task under the predicate-given setting as a sequence-to-sequence task like machine translation. Given a predicate, its SRL structure is converted into a token sequence. 
Their approach achieves competitive performance by using BART  \citep{bart-2020}.

Concurrently, \citet{zhang-2022-srlasdep} cast span-based SRL as \emph{a tree parsing approach}. 
Given a predicate, the word span corresponding to an argument is represented as latent trees. 
The sentence is encoded once without using predicate indicators, but each predicate require an independent decoding process.

\paragraph{Syntax-enhanced SRL.} 
Due to the close connection between syntax parsing and SRL, there has been a lot of works on syntax-enhanced SRL. 
\textcolor{black}{\citet{strubell-2018-linguistically} and \citet{zhou-2020-parsing} jointly handle  syntactic parsing and SRL under the multi-task learning framework. 
\citet{xia-2019-syntax} inject auto-parsed syntactic trees into SRL as extra features . 
}
In contrast, our work is a pure modeling study, and does not use external syntactic knowledge.

\paragraph{SDGP.}
In contrast to predicate-argument structure, SDGP  belongs to another category of semantic representation formalism, using  
word-based graphs to represent semantics of sentences \citep{oepen-2014-semeval, oepen-2015-semeval}. 
The specific forms include DM \citep{ivanova-etal-2012-dm}, PSD \citep{hajic-etal-2012-psd}, PAS \citep{miyao-2004-pas}, etc.

Straightforwardly, graph parsing is a mainstream approach for SDGP. 
\citet{dozat-2018-simpler} propose an efficient first-order graph parser to find an optimal graph from a fully connected graph. 
\citet{tukewei-2019-2osdp} extend the model of \citet{dozat-2018-simpler} by introducing second-order information. They compare two approximate high-order inference methods, i.e., mean filed variational inference and loopy belief propagation. 

\paragraph{Word-based graph parsing for word-based SRL.} 
As far as we know, \citet{li-2020-high} for the first time propose to treat word-based SRL as a SGDP task. 
Since arguments correspond to single words in  word-based SRL, the two tasks are very similar. 
They employ the SGDP model of  \citet{tukewei-2019-2osdp} straightforwardly. 
Moreover, their study focuses on the  \emph{predicate-given} setting.
First, they use a separate sequence labeling model to predict predicates. The SDGP model is then applied to recognize arguments.

\section{Proposed Graph Schemata}
\begin{figure}[tb]
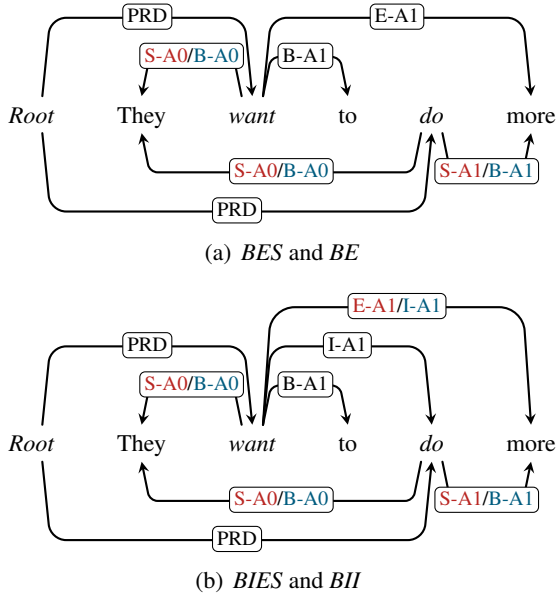

    \centering
    \subfigure[\emph{BES} and \emph{BE}]{
    \scalebox{1}{
    \begin{minipage}[t]{1\columnwidth}
            \centering
            \begin{dependency}[arc angle=80]
                \begin{deptext}[row sep=0.2cm, column sep=.65cm, font=\small]
                            \emph{Root} \& They \& \emph{want} \& to \& \emph{do} \& more \\
                \end{deptext}
                \depedge[edge style={black,thick}]{1}{3}{PRD}
                \depedge[edge style={black,thick}]{3}{2}{\red{S-A0}/\blue{B-A0}}
                \depedge[edge style={black,thick}]{3}{4}{B-A1}
                \depedge[edge style={black,thick}, edge height=6ex]{3}{6}{E-A1}
                \depedge[edge below,edge style={black,thick},edge height=6ex]{1}{5}{PRD}
                \depedge[edge below,edge style={black,thick},edge height=3ex]{5}{2}{\red{S-A0}/\blue{B-A0}}
                 \depedge[edge below,edge style={black,thick},edge height=3ex]{5}{6}{\red{S-A1}/\blue{B-A1}}
                
            \end{dependency}
        \end{minipage}%
        \label{fig:BE and BES}
    }
    }
    
    \subfigure[\emph{BIES} and \emph{BII}]{
    \scalebox{1}{
    \begin{minipage}[t]{1\columnwidth}
            \centering
            \begin{dependency}[arc angle=80]
                \begin{deptext}[row sep=0.05cm, column sep=.65cm, font=\small]
                        \emph{Root} \& They \& \emph{want} \& to \& \emph{do} \& more \\
                \end{deptext}
                \depedge[edge style={black,thick}]{1}{3}{PRD}
                \depedge[edge style={black,thick}]{3}{2}{\red{S-A0}/\blue{B-A0}}
                \depedge[edge style={black,thick}]{3}{4}{B-A1}
                \depedge[edge style={black,thick}]{3}{5}{I-A1}
                \depedge[edge style={black,thick}]{3}{6}{\red{E-A1}/\blue{I-A1}}
                \depedge[edge below,edge style={black,thick},edge height=6ex]{1}{5}{PRD}
                \depedge[edge below,edge style={black,thick},edge height=3ex]{5}{2}{\red{S-A0}/\blue{B-A0}}
                \depedge[edge below,edge style={black,thick},edge height=3ex]{5}{6}{\red{S-A1}/\blue{B-A1}}
            \end{dependency}
        \end{minipage}%
        }
        \label{fig:BII and BIES}
    }
    \caption{Proposed four different schemata. 
    Labels in black are the shared part. Red and blue labels belong to \emph{BES}, \emph{BIES} and \emph{BE}, \emph{BII} respectively. 
    }
    \label{fig:schemas}
\end{figure}

This work proposes to cast end-to-end span-based SRL as a  word-based graph parsing task. 
The key challenge is  %
to design a suitable graphical schema so that all predicates and their span-based arguments can be represented simultaneously in one graph without ambiguity.
And the graph can be transformed to its corresponding SRL structure without performance loss. 

\subsection{SRL-to-Graph Transformation}
We design four different schemata for transforming span-based SRL structures into word-based graphs. 
The basic idea is linking words in an  argument to the corresponding predicate,  and labeling the edges according to both semantic role labels  and word positions in the argument. 

Specifically, we add a pseudo ``\emph{Root}'' node at the beginning of the sentence and link all the predicates to it with ``\texttt{PRD}'' as the edge label.  
This allows our model to simultaneously predict predicates and arguments in an end-to-end manner.

Borrowing ideas from research on CWS and NER, we propose and investigate two strategies for attaching  argument words 
to corresponding predicates, i.e., \emph{boundary-attach} and \emph{all-attach}. %
The {boundary-attach} strategy connects only the start and end words of an argument to its predicate word, while the 
{all-attach} strategy connects all words of an argument to the predicate word. 
For each strategy, we design two concrete schemata, as follows.

\paragraph{Boundary-attach: \emph{BES} and \emph{BE}.}
Figure \ref{fig:BE and BES} shows the two schemata. 
When an argument contains multiple words, we attach only the start and end words to its corresponding predicates, using ``\texttt{B-$r$}'' and ``\texttt{E-$r$}'' as the edge labels, where  
$r$ is the original semantic role label.  
As shown in Figure \ref{fig:BE and BES}, the two schemata handle the argument ``to do more'' in the same way.

When an argument corresponds to a single word, for example, the argument ``They'',  the \emph{BE} schema simply uses ``\texttt{B-$r$}'' as the label, while the \emph{BES} schema uses ``\texttt{S-$r$}'' to make a distinction. Our experiments show that such distinction consistently improves performance.

\paragraph{All-attach: \emph{BIES} and \emph{BII}.} 
Figure \ref{fig:BII and BIES} shows the two schemata. 
Each word in an argument is attached to its corresponding predicate. 
In the \emph{BII} schema, the first word is labeled as ``\texttt{B-$r$}'', and the following words, if any, are labeled as ``\texttt{I-$r$}'', where  
the prefix ``\texttt{I-}'' means being inside an argument. 

Analogous to \emph{BES}, 
\emph{BIES} further distinguishes the end word in an argument using ``\texttt{E-$r$}'', and single-word arguments using ``\texttt{S-$r$}''.  %

In fact, there is another variant schema that belongs to the all-attach category, which is \emph{BIS}. Due to space limitation, we do not introduce  it in detail since our preliminary experiments show its performance lags behind the best schema by a large margin.

\subsection{Graph-to-SRL Recovery} \label{sec:graph-recover} 
In the evaluation stage, given an input sentence, our graph parsing model outputs an optimal graph  according to the underlying schema. 
Then, the job is to recover SRL structure. 
If the output graph is legal (i.e., without label conflicts), the recovery is quite straightforward.  
Taking the \emph{BES} schema for example, all children nodes (words) of the pseudo ``\emph{Root}'' are treated as predicates.
Then, for each predicate, we recover all its arguments based on the edge labels. An argument corresponds to either a paired labels, such as ``\texttt{B-A0}'' and ``\texttt{E-A0}'', or a single label such as ``\texttt{S-A0}''.

Unfortunately, it is quite complex to guarantee legality of output graphs.  
To handle this issue, we propose a simple yet effective post-processing recipe based on constrained Viterbi decoding in Section \ref{sec:vtb}.

\section{Model}
Based on our designed graphical schema, we can address span-based SRL as a word-based graph parsing task.
Following \citet{dozat-2018-simpler} and \citet{tukewei-2019-2osdp}, the framework of our model consists of two stages:
1) predicting all edges and 2) assigning labels for edges.
\begin{figure}[tb!]
    \centering
    \scalebox{0.65}{
    \begin{tikzpicture}[
        connect/.style={
                rounded corners=4pt,
                semithick,
                draw=black!80
            },
        arrow/.style={
                arrows = {-Straight Barb[length=0.5mm]},
                shorten >= 2pt,
                shorten <= 1.5pt,
                thin
            },
        inner arrow/.style={
                arrows = {-Straight Barb[length=0.4mm]},
                shorten >= 2pt,
                shorten <= 2pt,
                thin,
                draw=black!50
            },
        input/.style={
                rectangle,
                rounded corners=1mm,
                thin,
                dashed,
                draw=none,
                minimum width=3.5cm,
                minimum height=0.6cm,
            },
        share/.style={
                minimum height=0.5cm,
                  fill=orange!10,
                draw={rgb,255:red,225; green,209; blue,183},
                rounded corners=2mm,
            },
        chart/.style={
                circle,
                minimum size=3mm,
                draw=white,
                fill={rgb,255:red,189; green,215; blue,237},
                inner sep=0pt
        },
        chart1/.style={
                circle,
                minimum size=3mm,
                draw=white,
                fill={rgb,255:red,30; green,78; blue,120},
                inner sep=0pt
        },
        chart2/.style={
                circle,
                minimum size=3mm,
                draw=white,
                fill={rgb,255:red,205; green,223; blue,230},
                inner sep=0pt
        },
        chart3/.style={
                circle,
                minimum size=3mm,
                draw=white,
                fill={rgb,255:red,68; green,113; blue,199},
                inner sep=0pt
        },
        vilayer/.style={
                trapezium, 
                draw, 
                trapezium left angle=60, 
                trapezium right angle=120,
                rounded corners=1mm,
        },
        task4/.style={
                minimum height=0.5cm,
                draw=black,
                rounded corners=2mm,
                fill={rgb,255:red,240; green,249; blue,244},
            },
        task3/.style={
                minimum height=0.5cm,
                draw=white,
                rounded corners=2mm,
            },
        task2/.style={
                minimum height=0.5cm,
                fill={rgb,255:red,204; green,223; blue,230},
                draw={rgb,255:red,151; green,181; blue,191},
                rounded corners=2mm,
            },
        label/.style={
                inner sep=0.5mm,
                fill=white,
                minimum height=0.5cm,
            },
        task1/.style={
                minimum height=0.5cm,
                fill={rgb,255:red,253; green,210; blue,191},
                draw={rgb,255:red,233; green,133; blue,128},
                rounded corners=2mm
            },
        inner lstm/.style={
                fill=white,
                rectangle,
                rounded corners=1mm,
                semithick,
                draw=black!50,
                fill opacity=0.8
            },
        cell/.style={
                inner sep=2mm,
                rectangle,
                rounded corners=1mm,
                semithick,
                draw=black!50,
            },
        ocell/.style={
                solid,
                minimum height=0.5cm,
                rectangle,
                rounded corners=1mm,
                thick,
            },
        dep arrow/.style={
        arrows = {-Latex[round,open,length=8pt,width=6pt]},
        shorten >= 2pt,
        shorten <= 1.5pt,
        thick
        }
        ]
        \centering
        \node [input, inner sep=1pt] [minimum width=4.8cm] (input) at (0, 0) {$\ldots\;\; \mathbf{w}_i \;\;\ldots\;\; \mathbf{w}_k \;\;\ldots\;\; \mathbf{w}_j \;\;\ldots$};
        \node [inner sep=0mm] at ($(input.south) + (0, -0.5cm)$) {\textbf{}};
        \node [inner sep=0] (EmbedCat) at ($(input.north)$) {};
    
        \node [share, ocell] [minimum width=7cm, minimum height=0.675cm, anchor=south] (lstm) at ($(input.north) + (0, 0.3cm)$) {\scriptsize BiLSTM$\, \times \, 3$ \tiny or \scriptsize PLM};

        \draw [arrow, connect] ($(EmbedCat.north) + (0cm, -0.15cm)$) -- ($(lstm.south) + (0cm, 0)$);
    
        \foreach \x in {0, ..., 2}{
                \tikzmath{
                    integer \nx;
                    \nx = \x + 1;
                }
            };
        \node [share, fill=none, draw=gray, densely dashed] [minimum width=7cm, minimum height=0.775cm, anchor=south] (share-mlp) at ($(lstm.north) + (0cm, 0.455cm)$) {};
        \node [task2, ocell] [minimum width=1.25cm, minimum height=0.585cm, anchor=south west, fill opacity=0.85] (mlp-1-1-f) at ($(lstm.north west) + (0.1cm, 0.55cm)$) {};
        \node [task2, ocell] [minimum width=1.25cm, minimum height=0.585cm, anchor=south west, fill opacity=0.85] (mlp-1-1-b) at ($(lstm.north west) + (1.47cm, 0.55cm)$) {};
        \node [anchor=base] at  ($(mlp-1-1-f.south) + (0, 0.2cm)$)  {\scriptsize $\mathbf{MLP}^{h}$};
        \node [anchor=base] at  ($(mlp-1-1-b.south) + (0, 0.2cm)$)  {\scriptsize $\mathbf{MLP}^{m}$};

        \node [task2, ocell, fill=none, draw=none] [minimum width=1.75cm, minimum height=0.40cm, anchor=south east, fill opacity=0.5, draw opacity=0.6] (mlp-1-2-b) at ($(mlp-1-1-b.north east) + (0, 0.5cm)$) {};
    
        \node [task2, ocell, anchor=south, fill opacity=0.85, minimum height=1.29cm, minimum width=1.29cm, align=center] (mlp-1-2-f) at ($(mlp-1-1-f.north)!0.5!(mlp-1-1-b.north) + (0, 0.73cm)$) {\scriptsize Biaffine};
    
        \draw [arrow, connect] ($(mlp-1-1-b.north) + (0, 0.06cm)$) -- ($(mlp-1-1-b.north) + (0, 0.35cm)$) -- ($(mlp-1-2-f.south) + (0.2, 0)$);
        \draw [arrow, connect] ($(mlp-1-1-f.north) + (0, 0.06cm)$) -- ($(mlp-1-1-f.north) + (0, 0.35cm)$) -- ($(mlp-1-2-f.south) + (-0.2, 0)$);
    
        \node [task1, ocell] [minimum width=1.25cm, minimum height=0.585cm, anchor=south east, fill opacity=0.85] (mlp-2-1-f) at ($(lstm.north east) + (-2.85cm, 0.55cm)$) {};
        \node [task1, ocell] [minimum width=1.25cm, minimum height=0.585cm, anchor=south east, fill opacity=0.85] (mlp-2-1-b) at ($(lstm.north east) + (-1.48cm, 0.55cm)$) {};
        \node [task1, ocell] [minimum width=1.25cm, minimum height=0.585cm, anchor=south east, fill opacity=0.85] (mlp-2-1-bb) at ($(lstm.north east) + (-0.1cm, 0.55cm)$) {};
        \node [anchor=base] at  ($(mlp-2-1-f.south) + (0, 0.2cm)$)  {\scriptsize $\mathbf{MLP}^{h''}$};
        \node [anchor=base] at  ($(mlp-2-1-b.south) + (0, 0.2cm)$)  {\scriptsize $\mathbf{MLP}^{g}$};
        \node [anchor=base] at  ($(mlp-2-1-bb.south) + (0, 0.2cm)$)  {\scriptsize $\mathbf{MLP}^{m''}$};
        
        \node [task1, ocell, anchor=south, fill opacity=0.85, minimum height=1.29cm, minimum width=1.29cm, align=center] (tria3) at ($(mlp-2-1-f.north)!0.5!(mlp-2-1-bb.north) + (0.2cm, 0.93cm)$) {};
        \node [task1, ocell, anchor=south, fill opacity=0.85, minimum height=1.29cm, minimum width=1.29cm, align=center] (tria2) at ($(mlp-2-1-f.north)!0.5!(mlp-2-1-bb.north) + (0.1cm, 0.83cm)$) {};
        \node [task1, ocell, anchor=south, fill opacity=0.85, minimum height=1.29cm, minimum width=1.29cm, align=center] (tria1) at ($(mlp-2-1-f.north)!0.5!(mlp-2-1-bb.north) + (0cm, 0.73cm)$) {\scriptsize{Triaffines}};

        \draw [arrow, connect, rounded corners=10pt] ($(mlp-2-1-f.north) + (0, 0.06cm)$) -- ($(mlp-2-1-f.north) + (0, 1cm)$) -- ($(mlp-2-1-b.north) + (-0.875, 0.5) + (0.23, 0.23) + (0, 0.645)$);
        \draw [arrow, connect] ($(mlp-2-1-b.north) + (0, 0.06cm)$) -- ($(mlp-2-1-b.north) + (0, 0.5cm) + (0, 0.23)$);
        \draw [arrow, connect, rounded corners=10pt] ($(mlp-2-1-bb.north) + (0, 0.06cm)$) -- ($(mlp-2-1-bb.north) + (0, 1cm)$) -- ($(mlp-2-1-b.north) + (0.415, 0.5) + (0.23, 0.23) + (0.2, 0.645)$);
        
        \draw [arrow, connect, rounded corners=1.2pt, shorten >= 2pt] ($(lstm.north) + (0cm, 0)$) -- ($(share-mlp.south) + (0cm, 0)$);
        \draw [arrow, connect, rounded corners=1.2pt, shorten >= 2pt] ($(lstm.north) + (-2cm, 0)$) -- ($(share-mlp.south) + (-2cm, 0)$);
        \draw [arrow, connect, rounded corners=1.2pt, shorten >= 2pt] ($(lstm.north) + (2cm, 0)$) -- ($(share-mlp.south) + (2cm, 0)$);
        \node[anchor=base] at ($(share-mlp.south) + (-2.3cm,-0.32cm) $) (label-h1) {$\mathbf{h}_{i}$};
        \node[anchor=base] at ($(share-mlp.south) + (-0.3cm,-0.32cm) $) (label-h2) {$\mathbf{h}_{k}$};
        \node[anchor=base] at ($(share-mlp.south) + (1.7cm,-0.32cm) $) (label-h53) {$\mathbf{h}_{j}$};

        \node[anchor=base] at ($(share-mlp.north) + (-3cm,0.15cm) $) (label-r1) {$\mathbf{r}_{i}^{h}$};
        \node[anchor=base] at ($(share-mlp.north) + (-1.1cm,0.15cm) $) (label-r2) {$\mathbf{r}_{j}^{m}$};
        \node[anchor=base] at ($(share-mlp.north) + (-0.2cm,0.15cm) $) (label-r3) {$\mathbf{r}_{i}^{h''}$};
        \node[anchor=base] at ($(share-mlp.north) + (1.8cm,0.15cm) $) (label-r4) {$\mathbf{r}_{k}^{g}$};
        \node[anchor=base] at ($(share-mlp.north) + (3.1cm,0.15cm) $) (label-r5) {$\mathbf{r}_{j}^{m''}$};

        \node[anchor=base] at ($(mlp-1-2-f.north) + (0cm,0.5cm) $) (label-biaffine) {$\mathrm{s}(i,j)$};
        \node[anchor=base] at ($(tria1.north -| mlp-2-1-b) + (0,0.5cm) $) (label-triaffine) {$\mathrm{s}_{\ast}(i,k,j)$};
    
        \node [task3, ocell, anchor=south, fill opacity=0.85, minimum height=1.29cm, minimum width=1.29cm, align=center] (mfvi1) at ($(label-biaffine.north)!0.5!(label-triaffine.north) + (0cm, 0.3cm)$) {};
        
        \node[task4, minimum height=1.4cm, minimum width=1.4cm, anchor=south] (rec3) at ($(label-biaffine.north)!0.5!(label-triaffine.north) + (0.3cm, 0.6cm)$) {};
        \foreach \x in {1,...,4}{
                    \foreach \y in {1,...,4}{
                        \node[chart] (c_\x_\y) at (${\x-4}*(0.33, 0)+{\y-4}*(0, 0.33)+(rec3.north east)+(-0.22,-0.22)$) {};
                    }
                }
        \begin{scope}
                \node[chart2] at (c_2_4) {};
                \node[chart3] at (c_1_2) {};
                \node[chart2] at (c_2_3) {};
                \node[chart1] at (c_1_4) {};
                \node[chart2] at (c_3_2) {};
                \node[chart3] at (c_4_1) {};
                \node[chart2] at (c_4_3) {};
                \node[chart1] at (c_2_1) {};
        \end{scope}
        
        \node[task4, minimum height=1.4cm, minimum width=1.4cm, anchor=south] (rec2) at ($(label-biaffine.north)!0.5!(label-triaffine.north) + (0.15cm, 0.45cm)$) {};
        \foreach \x in {1,...,4}{
                    \foreach \y in {1,...,4}{
                        \node[chart] (b_\x_\y) at (${\x-4}*(0.33, 0)+{\y-4}*(0, 0.33)+(rec2.north east)+(-0.22,-0.22)$) {};
                    }
                }
         \begin{scope}
                \node[chart1] at (b_2_4) {};
                \node[chart1] at (b_1_2) {};
                \node[chart2] at (b_2_3) {};
                \node[chart2] at (b_1_4) {};
                \node[chart2] at (b_3_2) {};
                \node[chart2] at (b_4_1) {};
                \node[chart3] at (b_4_3) {};
                \node[chart3] at (b_2_1) {};
        \end{scope}
        
        \node[task4, minimum height=1.4cm, minimum width=1.4cm, anchor=south] (rec1) at ($(label-biaffine.north)!0.5!(label-triaffine.north) + (0cm, 0.3cm)$) {};
        
        \foreach \x in {1,...,4}{
                    \foreach \y in {1,...,4}{
                        \node[chart] (a_\x_\y) at (${\x-4}*(0.33, 0)+{\y-4}*(0, 0.33)+(rec1.north east)+(-0.22,-0.22)$) {};
                    }
                }
                
        \begin{scope}
                \node[chart1] at (a_2_4) {};
                \node[chart1] at (a_1_2) {};
                \node[chart2] at (a_2_3) {};
                \node[chart2] at (a_1_4) {};
                \node[chart2] at (a_3_2) {};
                \node[chart2] at (a_4_1) {};
                \node[chart3] at (a_4_3) {};
                \node[chart3] at (a_2_1) {};
        \end{scope}
        
        \draw [arrow, connect] ($(mlp-1-1-b.north) + (0, 0.06cm)$) -- ($(mlp-1-1-b.north) + (0, 0.35cm)$) -- ($(mlp-1-2-f.south) + (0.2, 0)$);
        \draw [arrow, connect] ($(mlp-1-1-f.north) + (0, 0.06cm)$) -- ($(mlp-1-1-f.north) + (0, 0.35cm)$) -- ($(mlp-1-2-f.south) + (-0.2, 0)$);
        
        \node[anchor=east] at ($(rec1.west) + (0.0cm,0.15cm) $) (label-mfvi) {\scriptsize{MFVI layers}};
        
        \draw [arrow, connect, rounded corners=10pt] ($(label-biaffine.north) + (0, -0.01cm)$) -- ($(label-biaffine.north) + (0, 0.42cm)$) -- ($(rec1.west) + (0, -0.15)$);
        \draw [arrow, connect, rounded corners=10pt] ($(label-triaffine.north) + (0, -0.01cm)$) -- ($(label-triaffine.north) + (0, 0.42cm)$) -- ($(rec1.east) + (0.12, -0.15)$);
        
        \draw [arrow, connect, rounded corners=1.2pt, shorten >= 2pt] ($(label-biaffine.south) + (0, -0.3cm) $) -- ($(label-biaffine.south) + (0, 0.15cm) $);
        \draw [arrow, connect, rounded corners=1.2pt, shorten >= 2pt, line cap=round] ($(label-triaffine.south) + (0, -0.3cm) $) -- ($(label-triaffine.south) + (0, 0.15cm) $);
        
        \node[anchor=south] at ($(rec2.north) + (0.0cm,0.3cm) $) (Q) {$Q_{i,j}^{T}$};
        \draw [arrow, connect, rounded corners=1.2pt, shorten >= 2pt] ($(Q.south) + (0, -0.3cm) $) -- ($(Q.south) + (0, 0.15cm) $);
    \end{tikzpicture}}
    \caption{Illustration of our model. $\mathrm{s}_{\ast}(i,k,j)$ corresponds to the second-order scores, where $\ast \in \{sib, cop, grd\}$.}
    \label{fig:model}
\end{figure}
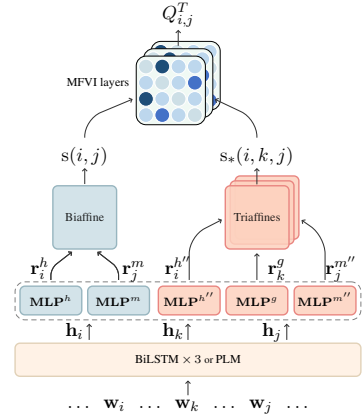

\subsection{Encoder}
\paragraph{BiLSTM.}
Under the setting without PLMs, we use BiLSTM as our encoder.
The input of the $i$-th word $w_i$ is the concatenation of word embedding $\mathbf{e}^{word}_i$, lemma embedding $\mathbf{e}^{lemma}_i$, and charLSTM representation vector:
\begin{equation} \label{equation:token-representation}
\mathbf{x}_i = \mathbf{e}^{word}_i \oplus \mathbf{e}^{lemma}_i \oplus \mathbf{e}^{char}_i
\end{equation}
where $\mathbf{e}^{char}_i$ is the output vector of a one-layer BiLSTM that encodes the character sequence \citep{lample-2016-neural}.
Then, %
a three-layer BiLSTM encoder produces a context-aware vector representation for each word.
\begin{equation} \label{equation:BiLSTM}
\mathbf{h}_i = \mathbf{f}_{i} \oplus \mathbf{b}_{i}
\end{equation}
where $\mathbf{f}_{i}$ and $\mathbf{b}_{i}$ respectively denote the output vectors of top-layer forward and backward LSTMs for $w_i$.

\paragraph{PLM.}
Under the setting with PLMs, we adopt ELMo \citep{peters-etal-2018-elmo} and BERT \citep{devlin-2018-bert} to get contextual word representation to boost the performance of our model.
\begin{equation} \label{equation:PLMs}
\mathbf{h}_i = \mathbf{PLM}(w_i)
\end{equation}
\textcolor{black}{Concretely, we use the outputs of all three layers in ELMO, and of the top four layers in BERT, and then apply weighted sum to obtain the final output vector for each token.
}

\subsection{Edge prediction}
In SDGP, the prediction of edge is treated as a binary 0/1 classification task, where 1 means that there exists an edge between the given word pair and 0 otherwise.
Here, for each edge $i \rightarrow j$
\footnote{For convenience, we abbreviate the edge $i \rightarrow j$ as $(i,j)$ in the remaining part of the paper.}
, we need to compute the logit score $\mathrm{logit}_{ij}$.
Then we can get the probability of existence of each edge $Q_{ij}$ by applying the Sigmoid function.
During inference, the edges that have $Q_{ij} > 0.5$ will be retained.

\begin{figure}[tb]
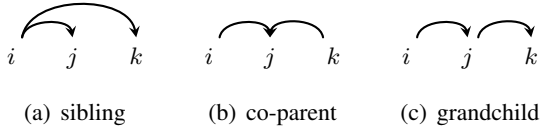

    \centering
    \subfigure[sibling]{
    \begin{minipage}[t]{0.3\columnwidth}
            \centering
            \begin{dependency}[arc edge, arc angle=80, text only label, label style={above}]
                \begin{deptext}[row sep=0.2cm, column sep=.5cm, font=\small]
                            $i$ \& $j$ \& $k$ \\
                \end{deptext}
                \depedge[edge style={black,thick}]{1}{2}{}
                \depedge[edge style={black,thick}]{1}{3}{}
            \end{dependency}
        \end{minipage}%
        \label{fig:sibling}
    }
    \subfigure[co-parent]{
    \begin{minipage}[t]{0.3\columnwidth}
            \centering
            \begin{dependency}[arc edge, arc angle=80, text only label, label style={above}]
                \begin{deptext}[row sep=0.4cm, column sep=.5cm, font=\small]
                            $i$ \& $j$ \& $k$ \\
                \end{deptext}
                \depedge[edge style={black,thick}]{1}{2}{}
                \depedge[edge style={black,thick}]{3}{2}{}
            \end{dependency}
        \end{minipage}%
        \label{fig:co-parent}
    }
    \subfigure[grandchild]{
    \begin{minipage}[t]{0.3\columnwidth}
            \centering
            \begin{dependency}[arc edge, arc angle=80, text only label, label style={above}]
                \begin{deptext}[row sep=0.4cm, column sep=.5cm, font=\small]
                             $i$ \& $j$ \& $k$\\
                \end{deptext}
                \depedge[edge style={black,thick}]{1}{2}{}
                \depedge[edge style={black,thick}]{2}{3}{}
            \end{dependency}
        \end{minipage}%
        \label{fig:grand}
    }
    
    \caption{Three types of second-order sub-trees.}
    \label{fig:second-order-stru}
\end{figure}

To facilitate computation and modeling, the first-order model of \citet{dozat-2018-simpler} makes a strong assumption that edges 
are mutually independent and thus it only considers the information between the current two words when computing logits.
However, in our case, the edges in the resulting graph usually have a strong correlation. For example, in our \emph{BE} schema, a ``\texttt{B-$*$}'' edge usually calls for a ``\texttt{E-$*$}'' edge, and vice-versa, to form a complete argument.
So, in this work, we extend first-order to second-order by adding three types of sub-trees, as shown in Figure \ref{fig:second-order-stru}. 
And we compute the logit by mean field variational inference (MFVI) following \citet{tukewei-2019-2osdp}. 

The logit comes from two parts. The first part is the first-order score $\mathrm{s}(i,j)$. 
We use two MLPs to get representation vectors of a word as a head or a modifier respectively, and then use a BiAffine to compute edges' first-order scores as follows:
\begin{equation} \label{eq:edge-scoring}
\begin{split}
\mathbf{r}_i^{\mathrm{h}}; \mathbf{r}_i^{\mathrm{m}} & =\mathrm{MLP}^{\mathrm{h}} \left( \mathbf{h}_i \right); \mathrm{MLP}^{\mathrm{m}} \left( \mathbf{h}_i \right) \\
\mathrm{s}(i,j) & =  \left[
\begin{array}{c}
  \mathbf{r}_{j}^{\mathrm{m}} \\
    1
\end{array}
\right]^{\top}
\mathbf{W} \mathbf{r}_{i}^{\mathrm{h}} \\
\end{split}
\end{equation}
where $\mathbf{W} \in \mathbb{R}^{(d+1) \times d}$.

The other part comes from second-order sub-trees. 
First, we use three new MLPs to get representations of each word for playing different roles in second-order sub-trees as follows: 
\begin{equation}
\label{mlp-high-order}
\mathbf{r}_i^{\mathrm{h''}}; \mathbf{r}_i^{\mathrm{m''}}; \mathbf{r}_i^{\mathrm{g}} =\mathrm{MLP}^{\mathrm{h''/m''/g}} \left( \mathbf{h}_i \right)
\end{equation}
where $\mathbf{r}_i^{h''}$, $\mathbf{r}_i^{m''}$, and $\mathbf{r}_i^{g}$ denote the representation vectors of $w_i$ as head, modifier, and grandchild respectively.
Then, a TriAffine scorer \citep{zhang-2020-efficient} taking the three vectors as input is applied to compute the score of the corresponding second-order structure:
\begin{equation} \label{equation:TriAffine}
\mathrm{TriAFF}(\mathbf{v}_{1}, \mathbf{v}_{2}, \mathbf{v}_{3}) =
\left[
\begin{array}{c}
\mathbf{v}_{3} \\
1
\end{array}
\right]^{\top}
{\mathbf{v}_{1}}^{\top}
\mathbf{W'}
\left[
\begin{array}{c}
\mathbf{v}_{2} \\
1
\end{array}
\right]
\end{equation}
where $\mathbf{W'} \in \mathbb{R}^{(d'+1) \times d' \times (d'+1)}$ and $\mathbf{v}_{1,2,3} \in \mathbb{R}^{d'}$.
Finally, scores of the three types of sub-trees can be computed as follows respectively:
\begin{align}
    \mathrm{s}_{sib}(i,j,k) &= \mathrm{TriAFF}_{sib}(\mathbf{r}_i^{\mathrm{h''}}, \mathbf{r}_j^{\mathrm{m''}}, \mathbf{r}_k^{\mathrm{m''}})\\
    \mathrm{s}_{cop}(i,j,k) &= \mathrm{TriAFF}_{cop}(\mathbf{r}_i^{\mathrm{h''}}, \mathbf{r}_j^{\mathrm{m''}}, \mathbf{r}_k^{\mathrm{h''}})\\
    \mathrm{s}_{grd}(i,j,k) &= \mathrm{TriAFF}_{grd}( \mathbf{r}_i^{\mathrm{h''}}, \mathbf{r}_j^{\mathrm{m''}}, \mathbf{r}_k^{\mathrm{g}})
\end{align}
It should be noted that for symmetrical sibling sub-trees and co-parent sub-trees, we compute their corresponding scores only once, i.e.,  $\mathrm{s}_{sib}(i,j,k)=\mathrm{s}_{sib}(i,k,j)$ and  $\mathrm{s}_{cop}(i,j,k)=\mathrm{s}_{cop}(k,j,i)$.

For a given edge $(i,j)$, MFVI aggregates the final $\mathrm{logit}_{ij}^{T}$ and $Q_{ij}^{T}$ from the corresponding first-order score and second-order scores iteratively as follows:
\begin{equation}\label{equation:mfvi-update}
    \begin{gathered}
        \begin{split}
            \mathcal{M}_{ij}^{t-1} =&\sum_{k \neq i,j} Q_{ik}^{t-1}\mathrm{s}_{sib}(i,j,k) \\
            &+Q_{kj}^{t-1}\mathrm{s}_{cop}(i,j,k) \\
            &+Q_{jk}^{t-1}\mathrm{s}_{grd}(i,j,k)
        \end{split} \\
        \begin{aligned} 
            \mathrm{logit}_{ij}^{t} & = \mathrm{s}(i,j) + \mathcal{M}_{ij}^{t-1} \\
            Q_{ij}^{t} & =
            \sigma(\mathrm{logit}_{ij}^{t}) 
        \end{aligned} 
    \end{gathered}
\end{equation}
where $t \in [1,T]$ is the iteration number. $\mathcal{M}_{ij}$ is an intermediate variable that stores message from second-order sub-tree scores. $Q_{ij}^{0}$ is initialized by applying Sigmoid on $\mathrm{s}(i,j)$. Through $T$ times of update, we get the final $\mathrm{logit}_{ij}^{T}$ and probability $Q_{ij}^{T}$.

\subsection{Label prediction}
Similar to edge scoring, we use two extra MLPs and a set of Biaffines to compute the label scores: 
\begin{equation}\label{eq:label-scoring}
\begin{split}
\mathbf{r}_i^{\texttt{h'}}; \mathbf{r}_i^{\texttt{m'}} & =\mathrm{MLP}^{\mathrm{h'}} \left( \mathbf{h}_i \right); \mathrm{MLP}^{\mathrm{m'}} \left( \mathbf{h}_i \right) \\
\mathrm{s}(i,j,\ell) &=  \left[
\begin{array}{c}
  \mathbf{r}_{j}^{\mathrm{m'}} \\
    1
\end{array}
\right]^{\top}
\mathbf{W}^{\texttt{label}}_{\ell} \left[\begin{array}{c}
  \mathbf{r}_{i}^{\mathrm{h'}} \\
    1
\end{array}\right] \\
\mathrm{p}(\ell|i,j) &= \cfrac{\exp{(\mathrm{s}(i,j,\ell))}}{\sum_{\ell^{'} \in \mathcal{L}}^{}{\exp{(\mathrm{s}(i,j,\ell^{'}))}}}
\end{split}
\end{equation}
where $\mathrm{s}(i,j,\ell)$ is the score of the label $\ell$ for the edge $(i,j)$.  $\mathrm{p}(\ell|i,j)$ is the probability after softmax over all labels.
Each label has its own Biaffine parameters 
$\mathbf{W}^{\texttt{label}}_{\ell} \in \mathbb{R}^{(d+1) \times (d+1)}$.

\subsection{Training} \label{sec:training}
The loss of our system comes from both edge and label prediction modules. 
Given one sentence $X$ and its gold graph $G$, the fully connected graph of $X$ is denoted as $C$. 
\begin{equation}\label{eq:xx}
    \begin{aligned}
        \emph{L}_{e}(\theta) &= -\sum_{\mathclap{(i,j)\in G}}{\log{Q_{ij}^{T}}} - \sum_{\mathclap{(i,j)\in C\backslash G}}{\log{(1-Q_{ij}^{T}})}\\
        \emph{L}_{l}(\theta) &=
        -\sum_{\mathclap{(i,j)\in G}}\log{\mathrm{p}(\hat{\ell}|i,j)}
    \end{aligned}
\end{equation}
where $\theta$ denotes model parameters;
$C\backslash G$ is the set of incorrect edges; 
$\hat{\ell}$ is the gold label of edge. 
The loss of the final model is the weighted sum of the two parts: 
\begin{equation}\label{eq:loss-func}
    \emph{L}(\theta) = \lambda \emph{L}_{l}(\theta) + (1-\lambda)\emph{L}_{e}(\theta)
\end{equation}
where $\lambda = 0.06$ in our model.

\section{Conflict resolution} \label{sec:vtb}
During inference, we first use the edge prediction module to build the graph skeleton, and then use the label prediction module to assign labels to predicted edges.
After that, we use a simple procedure to check whether the generated graph is legal. 
Concretely, for each predicate, we scan the edges of the predicate from left to right.
For example, in the \emph{BES} schema, a ``\texttt{B-$\ast$}'' edge must be followed by a ``\texttt{E-$\ast$}'' edge; ``\texttt{S-$\ast$}'' edge and ``\texttt{E-$\ast$}'' can be followed by a ``\texttt{B-$\ast$}'' edge or ``\texttt{S-$\ast$}'' edge.
If the generated graph is legal, we can directly recover the corresponding SRL structure through Graph-to-SRL procedure described in \ref{sec:graph-recover}.

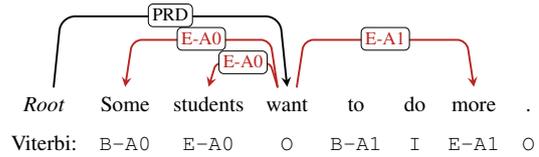
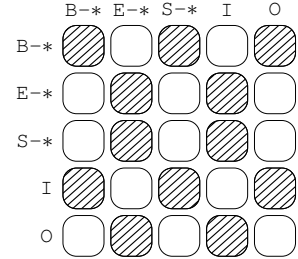
\begin{figure}[tb!]
    \centering
    \subfigure[A conflicting example in \emph{BES}. Edges in red cause conflicts, and the sequence below is the corrected sequence via our constrained Viterbi.]{
    \scalebox{0.9}{
    \begin{minipage}[t]{1\linewidth}
            \centering
            \begin{dependency}[]
                \begin{deptext}[column sep=.15cm, row sep=0.1cm, font=\small]
                            \emph{Root} \& Some \& students \& want \& to \& do \& more \& . \\
                            Viterbi:
                            \& \textcolor{black}{\texttt{B-A0}} \& \textcolor{black}{\texttt{E-A0}} \& \textcolor{black}{\texttt{O}} \& \textcolor{black}{\texttt{B-A1}} \& \textcolor{black}{\texttt{I}} \& \textcolor{black}{\texttt{E-A1}} \& \textcolor{black}{\texttt{O}} \\
                \end{deptext}
                \depedge[edge style={black,thick}, edge height=6ex]{1}{4}{\color{black}PRD}
                \depedge[edge style={brickred,thick}, edge height=4ex]{4}{2}{\red{E-A0}}
                \depedge[edge style={brickred,thick}, edge height=2ex]{4}{3}{\red{E-A0}}
                \depedge[edge style={brickred,thick}, edge height=4ex]{4}{7}{\red{E-A1}}
            \end{dependency}
        \end{minipage}}%
        \label{fig:conflict-example}
    }
    \subfigure[The transition matrix of \emph{BES}. ]{
    \scalebox{0.9}{
    \begin{tikzpicture}[
    forbid_chart/.style={
            minimum size=0.6cm,
            rounded corners=2mm,
            draw=black,
            minimum width=0.6cm,
            minimum height=0.6cm,
            pattern=south west lines,
        },
    chart/.style={
            minimum size=0.6cm,
            rounded corners=2mm,
            draw=black,
            minimum width=0.6cm,
            minimum height=0.6cm,
    }
    ]
        \foreach \x in {1,...,5}{
                \foreach \y in {1,...,5}{
                    \node[chart] (a_\x_\y) at (${\x-4}*(0.7, 0)+{\y-4}*(0, 0.7)$) {};
                }
            }
        \begin{scope}
            \node[forbid_chart] at (a_1_2) {};
            \node[forbid_chart] at (a_1_5) {};
            \node[forbid_chart] at (a_2_1) {};
            \node[forbid_chart] at (a_2_3) {};
            \node[forbid_chart] at (a_2_4) {};
            \node[forbid_chart] at (a_3_2) {};
            \node[forbid_chart] at (a_3_5) {};
            \node[forbid_chart] at (a_4_1) {};
            \node[forbid_chart] at (a_4_3) {};
            \node[forbid_chart] at (a_4_4) {};
            \node[forbid_chart] at (a_5_2) {};
            \node[forbid_chart] at (a_5_5) {};
        \end{scope}

        \node[anchor=east] at (a_1_5.west) {\small \texttt{B-$\ast$}};
        \node[anchor=east] at (a_1_4.west) {\small \texttt{E-$\ast$}};
        \node[anchor=east] at (a_1_3.west) {\small \texttt{S-$\ast$}};
        \node[anchor=east] at (a_1_2.west) {\small \texttt{I}};
         \node[anchor=east] at (a_1_1.west) {\small \texttt{O}};
        
        \node[anchor=south] at (a_1_5.north) {\small \texttt{B-$\ast$}};
        \node[anchor=south] at (a_2_5.north) {\small \texttt{E-$\ast$}};
        \node[anchor=south] at (a_3_5.north) {\small \texttt{S-$\ast$}};
        \node[anchor=south] at (a_4_5.north) {\small \texttt{I}};
        \node[anchor=south] at (a_5_5.north) {\small \texttt{O}};
        
    \end{tikzpicture}}
    \label{fig:tran-matrix}
    }
    \caption{A conflicting example and the transition matrix in \emph{BES} schema. 
    The rows indicate the beginning of the transition and the columns indicate the ending.
    Cells with fence denote the prohibited transitions. \texttt{I} and \texttt{O} are two pseudo labels. 
    }
    \label{fig:comflict example and matrix}
\end{figure}

However, 
since the label prediction module handles each edge independently, 
the resulting graph may contain conflicts, 
 as shown in the upper part of Figure \ref{fig:conflict-example} \footnote{Here we only take the \emph{BES} schema as a representative for discussion, and others can be viewed in \S~\ref{app:other_matrix}.}. 
First, if two consecutive edges are both labeled as ``\texttt{E-$\ast$}'', such as the two ``\texttt{E-A0}'' edge, then it is impossible to recover the corresponding arguments. 
Another conflicting scene is when there exists a single outlier edge labeled as ``\texttt{B-$\ast$}'' or ``\texttt{E-$\ast$}'', such as ``\texttt{E-A1}'' edge in the figure. 

\paragraph{Constrained Viterbi.}
We propose to employ constrained decoding to handle conflicts. Concretely, when conflicts occur during recovering arguments for a predicate in the output graph, we re-label all words in the sentence for the predicate.
However, it is non-trivial to apply constrained Viterbi to our SDGP framework as a post-processing step. 

Here we use the \emph{BES} schema as an example, and the process for other schemata is similar. In the first  stage, $Q_{ij}^{T}$ means the probability that the edge appears in the final graph; while in the second stage, $\mathrm{p}(\ell|i,j)$ means the probability that the edge should be labeled as $\ell \in \mathcal{L}=\{\mathrm{\texttt{B-*}}, \mathrm{\texttt{E-*}}, \mathrm{\texttt{S-*}}\}$. We can see that $\mathcal{L}$ does not include ``\texttt{I}'' and ``\texttt{O}'',  meaning that the word is inside an argument or outside any arguments respectively. The two labels are indispensable for the sequence labeling procedure.   

To solve this issue, we add two pseudo labels  ``\texttt{O/I}'' into the label set, and redistribute the label probability distribution as follows. 
\begin{equation}\label{equation:emitting probability}
    \begin{aligned}
        \mathrm{p'}(\ell|i,j) &=Q_{ij}^{T} \cdot \mathrm{p}(\ell|i,j) \\
        \mathrm{p'}(\texttt{O}|i,j) &= \mathrm{p'}(\texttt{I}|i,j) = 1-Q_{ij}^{T} \\
    \end{aligned}
\end{equation}
where $\mathrm{p'}(\ell|i,j)$ is the probability for the normal label such as ``\texttt{B-A0}''. $\mathrm{p'}(\texttt{O}|i,j)$ and $\mathrm{p'}(\texttt{I}|i,j)$ share the same probability because
they both mean that there is no edge pointing to the word,
but ``\texttt{I}'' has an extra indication that there is an unpaired ``\texttt{B-$\ast$}'' in the left side.
Thus, we can solve the conflicts by controlling the transition matrix.

For example, as shown in Figure \ref{fig:tran-matrix}, 
we disallow transitions from ``\texttt{E-$\ast$}'' to ``\texttt{E-$\ast$}''. So, the ``Some'' and ``students'' are re-labeled as ``\texttt{B-A0}'' and ``\texttt{E-A0}''. And finally we get the correct argument span ``Some students'' with semantic role ``\texttt{A0}''.

\section{Experiments}

\paragraph{Data and evaluation.}
Experiments are conducted on CoNLL05 \citep{palmer-etal-2005-proposition} and larger-scale CoNLL12 \citep{pradhan-etal-2012-conll}, which are two widely used span-based SRL datasets. 
Following previous works on span-based SRL, we omit predicate sense prediction \cite{zhou-2015-end, he-etal-2017-deep}.
We use the official evaluation scripts\footnote{\url{http://www.cs.upc.edu/~srlconll/st05/st05.html}}.
We choose seeds randomly to run our model for 3 times and report the average results.

\paragraph{Hyper-parameter settings.} \label{app:parameter}
We employ 300-dimension English word embeddings from GloVe \citep{pennington-etal-2014-glove} for our experiments.
We adopt most hyper-parameters of the SDGP work of \citet{tukewei-2019-2osdp}, except that 
we reduce the dimension of Char-LSTM from 400 to 100 to save the memory, which only slightly influence performance. 
For experiments with PLMs, we adopt ELMo
\citep{peters-etal-2018-elmo} 
and BERT
\citep{devlin-2018-bert} as our encoder. 
Following most of previous works \citep{he-2018-jointly, xia-2019-syntax}, for ELMo, we froze its parameters during training.
For BERT, we fine-tune its parameters for 10 epochs.
The initial learning rate for models with and without BERT are  5e-5 and 1e-3 respectively.
The hyper-parameter $\lambda$ in the loss function (Eq. \ref{eq:loss-func}) is set to 0.06 in all experiments, based on preliminary experiment results.

\begin{table}[!tb]
\begin{small}
    \centering
    \setlength{\tabcolsep}{4.0pt}
    \begin{tabular}{l ccc ccc}
         \toprule
        \multirow{2}{*}{Schema} &
        \multicolumn{3}{c}{WSJ} & \multicolumn{3}{c}{Brown} \\
        \cmidrule(lr){2-4} \cmidrule(lr){5-7}
        & P & R & $\textrm{F}_1$ & P & R & $\textrm{F}_1$ \\
        \hline
        \emph{BES} & \bf{85.28} & 83.66 & \bf{84.46} & \bf74.10 & 70.76 & \bf72.39 \\
        \emph{BE} & 83.97 & 83.56 & 83.76 & 71.82 & 70.19 & 70.99 \\
        \emph{BIES} & 82.63 & \bf83.92 & 83.27 & 70.22 & \bf72.03 & 71.11 \\
        \emph{BII} & 81.65 & 83.44 & 82.54 & 67.72 & 70.74 & 69.20 \\
        \hline
        \multicolumn{7}{l}{\textbf{+BERT}} \\
        \emph{BES} & \bf87.15 & \bf88.44 & \bf87.79 & \bf79.44 & 80.85 & \bf80.14 \\
        \emph{BE} & 86.37 & 87.93 & 87.14 & 78.18 & 79.91 & 79.04 \\
        \emph{BIES} & 85.91 & 88.17 & 87.03 & 77.59 & \bf81.76 & 79.62 \\
        \emph{BII} & 85.31 & 87.57 & 86.43 & 76.90 & 81.03 & 78.91 \\
        \bottomrule
    \end{tabular}
    \caption{Results on CoNLL05 datasets with respect to proposed four schemata. The variation between the 3 runs on WSJ and Brown is about 0.1 and 0.2, respectively. And it varies little between different schemata. 
} 
    \label{table:4schema}
\end{small}
\end{table}
\subsection{Schema Comparison}
\paragraph{Overall results.}
First, to compare the proposed four schemata and find which one is better, we conduct experiments on CoNLL05 datasets under the \emph{end-to-end} setting.
Table \ref{table:4schema} shows results of different schemata.
First, by comparing the two different attaching strategy, i.e., all-attach (\emph{BII}, \emph{BIES}) and boundary-attach (\emph{BE}, \emph{BES}), we can find that the schemata resulted from boundary-attach have better P and $\textrm{F}_1$ results.
We think this may be because \emph{BII} and \emph{BIES} connect all words in arguments to predicates. So the final graph contains much more edges than that generated by \emph{BE} and \emph{BES}. Therefore, given two words, the corresponding model tends to build an edge between them, compared with not building an edge, 
resulting in a higher R but a lower P. 
Second, by comparing schemata that with and without ``\texttt{S-$r$}'', i.e., \emph{BII} vs. \emph{BIES} and \emph{BE} vs. \emph{BES}, we find that it is always better to use a separate ``\texttt{S-$r$}'' to label the edge corresponding to the single-word argument.
Therefore, from the overall point of view, we can get the conclusion that \emph{BES} > \emph{BE} > \emph{BIES} > \emph{BII}.

\paragraph{Performance regarding argument width.}
Here we define an argument's width as the number of words included.
As we know, different schemata have different attaching and labeling methods to represent arguments of the same width.
Therefore, analyzing the performance of different schemata on the same width argument will help us to deeply explore the advantages and disadvantages of schemata.

\begin{figure}[tb]
    \centering
    \includegraphics[width=1\columnwidth]{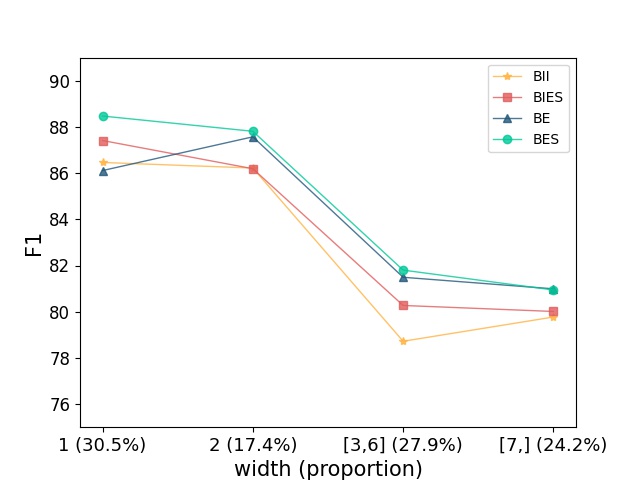}
    
    \caption{Analysis of the arguments with different width. The horizontal axis denotes the width of arguments and the proportion of arguments of the same width in the data set. The vertical axis denotes the $\mathrm{F}_1$ value.}
    \label{fig:analy_on_length}
\end{figure}

As shown in Figure \ref{fig:analy_on_length}, we divide arguments into four categories according to their width, and report $\mathrm{F}_1$ values for each category. The proportion of each category in the gold-standard data is also reported. 
First, we can see that \emph{BES} and \emph{BIES} perform much better on 1-width arguments. This further shows that it is necessary to use ``\texttt{S-$r$}'' alone to represent arguments of width 1.
Then, we can clearly find that \emph{BE} and \emph{BES} perform better than \emph{BII} and \emph{BIES} on arguments containing multiple words.
And we know that \emph{BE} and \emph{BES} are resulted from the boundary-attach strategy which pays more attention to boundary information. 
So, we may conclude that boundary information is more helpful to the recognition of multi-word arguments.

Through analyzing the performance of different schemata, we find that \emph{BES} is more suitable for converting span-based SRL into word-based graphs than other schemata.
So, the rest of the experiments are conducted in \emph{BES} schema.

\subsection{Efficiency}
\begin{table}[!tb]
\begin{small}
    \centering
    \begin{tabular}{l l r}
         \toprule
        Model & Type & Sents/sec \\
         \hline
         \citet{he-2018-jointly} & SGP & 44 \\
         \citet{strubell-2018-linguistically} & BIO-based & 45 \\
         \citet{zhang-2022-srlasdep} &  \multirow{2}{*}{TP} & 214 \\
         \citet{zhang-2022-srlasdep}$_{\mathrm{\texttt{BERT}}}$ &  & 113 \\
         Ours & \multirow{2}{*}{WGP} & \bf669 \\
         Ours$_{\mathrm{\texttt{BERT}}}$ & & 252 \\
        \bottomrule
    \end{tabular}
    \caption{Speed comparison on CoNLL05. 
    ``SGP'' and ``WGP''denote the span-based graph parsing and word-based graph parsing approach respectively; ``TP'' means the tree parsing approach.
    }
    \label{table:speed}
\end{small}
\end{table}
\begin{table*}[!tb]
\begin{small}
    \centering
    \setlength{\tabcolsep}{4.5pt}
    \begin{tabular}{l cccc ccc cccc}
         \toprule
        \multirow{2}{*}{Model} &
        \multicolumn{4}{c}{CoNLL05-WSJ} & \multicolumn{3}{c}{CoNLL05-Brown} & \multicolumn{4}{c}{CoNLL12} \\
        \cmidrule(lr){2-5} \cmidrule(lr){6-8} \cmidrule(lr){9-12}
        & Dev.$\textrm{F}_1$ & P & R & $\textrm{F}_1$ & P & R & $\textrm{F}_1$ & Dev.$\textrm{F}_1$ & P & R & $\textrm{F}_1$ \\
        \hline
        \multicolumn{12}{c}{\bf{{The end-to-end setting}}} \\
        \citet{he-etal-2017-deep}$^{\dagger}$ &
        80.3\textcolor{white}{0} &
        80.2\textcolor{white}{0} & 82.3\textcolor{white}{0} & 81.2\textcolor{white}{0} & 67.6\textcolor{white}{0} & 69.6\textcolor{white}{0} & 68.5\textcolor{white}{0} &
        75.5\textcolor{white}{0} &
        78.6\textcolor{white}{0} & 75.1\textcolor{white}{0} & 76.8\textcolor{white}{0} \\
        \citet{strubell-2018-linguistically}$^{\dagger}$ $^\ast$ & 81.72 & 81.77 & 83.28 & 82.51 & 68.58 & 70.10 & 69.33 & - & - & - & - \\
        \citet{he-2018-jointly}$^{\ddagger}$ &
        81.6\textcolor{white}{0} &
        81.2\textcolor{white}{0} & {83.9}\textcolor{white}{0} & 82.5\textcolor{white}{0} & 69.7\textcolor{white}{0}& {71.9}\textcolor{white}{0} & 70.8\textcolor{white}{0} &
        79.4\textcolor{white}{0} &
        79.4\textcolor{white}{0} & {80.1}\textcolor{white}{0} & 79.8\textcolor{white}{0}\\
        \citet{Li-2019-dependency}$^{\ddagger}$ & - & - & - & 83.0\textcolor{white}{0} & - & - & - & - & - & - & -\\
        \citet{zhou-2020-parsing} & 82.27 & - &- &- &- &- &- &- &- &- &- \\ 
        \citet{zhang-2022-srlasdep} & \bf83.91 & 83.26 & \bf86.20 & \bf84.71 & 70.70 & \bf74.16 & \bf72.39 & \bf81.16 & 79.27 & \bf83.24 & \bf81.21 \\
        Ours & 83.17 & \bf85.28 & 83.66 & 84.46 & \bf74.10 & 70.76 & \bf72.39 & 80.79 & \bf82.10 & 79.76 & 80.91 \\
        \hline
        \citet{strubell-2018-linguistically}$^{\dagger}$ $^\ast$ \hfill + ELMo & 84.73 & 83.86 & 85.98 & 84.91 & 73.01 & 75.61 & 74.31 & - & - & - & - \\
        \citet{he-2018-jointly}$^{\ddagger}$ \hfill + ELMo &
        85.3\textcolor{white}{0} &
        84.8\textcolor{white}{0} & 87.2\textcolor{white}{0} & 86.0\textcolor{white}{0} & 73.9\textcolor{white}{0} & 78.4\textcolor{white}{0} & 76.1\textcolor{white}{0} &
        83.0\textcolor{white}{0} &
        81.9\textcolor{white}{0} & 84.0\textcolor{white}{0} & 82.9\textcolor{white}{0} \\
        \citet{Li-2019-dependency}$^{\ddagger}$ \hfill + ELMo &
        - &
        85.2\textcolor{white}{0} & 87.5\textcolor{white}{0} & 86.3\textcolor{white}{0} & 74.7\textcolor{white}{0} & 78.1\textcolor{white}{0} & 76.4\textcolor{white}{0} &
        - &
        \bf{84.9}\textcolor{white}{0} & 81.4\textcolor{white}{0} & 83.1\textcolor{white}{0}\\
        Ours \hfill + ELMo & 85.72 & 86.19 & 86.91 & 86.55 & 76.57 & 77.77 & 77.17 & 83.72 & 83.53 & 83.56 & 83.54 \\
        \citet{zhang-2022-srlasdep} \hfill + BERT & \bf87.03 & 87.00 & \bf88.76 & \bf87.87 & 79.08 & \bf81.50 & \bf80.27 & \bf85.53 & \bf84.53 & \bf86.41 & \bf85.45 \\
        Ours \hfill + BERT & 86.79 & \bf87.15 & 88.44 & 87.79 & \bf79.44 & 80.85 & 80.14 & 84.74 & 83.91 & 85.61 & 84.75 \\
        \hline
        \hline
        \multicolumn{12}{c}{\bf{{The predicate-given setting}}} \\
        \citet{he-etal-2017-deep}$^{\dagger}$ & 81.6\textcolor{white}{0} & 83.1\textcolor{white}{0} & 83.0\textcolor{white}{0} & 83.1\textcolor{white}{0} & 72.9\textcolor{white}{0} & 71.4\textcolor{white}{0} & 72.1\textcolor{white}{0} & 81.5\textcolor{white}{0} & 81.7\textcolor{white}{0} & 81.6\textcolor{white}{0} & 81.7\textcolor{white}{0} \\
        \citet{strubell-2018-linguistically}$^{\dagger}$ & - & 84.70 & 84.24 & 84.47 & 73.89 & 72.39 & 73.13 & - & - &- & - \\
        \citet{he-2018-jointly}$^{\ddagger}$ & - & - & - & 83.9\textcolor{white}{0} & - & - & 73.7\textcolor{white}{0} & - & - & - & 82.1\textcolor{white}{0} \\
        \citet{tan-2018-deep}$^{\dagger}$ & 83.1\textcolor{white}{0} & 84.5\textcolor{white}{0} & 85.2\textcolor{white}{0} & 84.8\textcolor{white}{0} & 73.5\textcolor{white}{0} & 74.6\textcolor{white}{0} & 74.1\textcolor{white}{0} & 82.9\textcolor{white}{0} & 81.9\textcolor{white}{0} & 83.6\textcolor{white}{0} & 82.7\textcolor{white}{0} \\
        \citet{zhou-2020-parsing} & 83.16 &- &- &- &- &- &- &- &- &- &- \\
        \citet{zhang-comparing-spanextra} & 84.45 & 85.30 & 85.17 & 85.23 & 74.98 & 73.85 & 74.41 & 82.83 & 83.09 & 83.71 & 83.40\\
        \citet{zhang-2022-srlasdep} & \bf84.65 & 85.47 & \bf86.40 & \bf85.93 & 74.92 & \bf75.00 & 74.96 & 83.39 & 83.02 & \bf84.31 & 83.66 \\
        Ours & 84.39 & \bf87.01 & 84.36 & 85.66 & \bf77.86 & 72.53 & \bf75.10 & \bf83.83 & \bf85.74 & 82.95 & \bf84.32 \\
        \hline
        \citet{Li-2019-dependency}$^{\ddagger}$ \hfill + ELMo & - & 87.9\textcolor{white}{0} & 87.5\textcolor{white}{0} & 87.7\textcolor{white}{0} & 80.6\textcolor{white}{0} & 80.4\textcolor{white}{0} & 80.5\textcolor{white}{0} & - & 85.7\textcolor{white}{0} & 86.3\textcolor{white}{0} & 86.0\textcolor{white}{0} \\
        \citet{shi-2019-simple}$^{\dagger}$ \hfill + BERT & - & 88.6\textcolor{white}{0} & 89.0\textcolor{white}{0} & 88.8\textcolor{white}{0} & 81.9\textcolor{white}{0} & 82.1\textcolor{white}{0} & 82.0\textcolor{white}{0} & - & 85.9\textcolor{white}{0} & 87.0\textcolor{white}{0} & 86.5\textcolor{white}{0} \\
        \citet{jindal-2020-improved}$^{\dagger}$ \hfill + BERT & 87.1\textcolor{white}{0} & 87.7\textcolor{white}{0} & 88.0\textcolor{white}{0} & 87.9\textcolor{white}{0} & 80.3\textcolor{white}{0} & 80.1\textcolor{white}{0} & 80.2\textcolor{white}{0} & 86.6\textcolor{white}{0} & 86.3\textcolor{white}{0} & 86.8\textcolor{white}{0} & 86.6\textcolor{white}{0} \\
        \citet{zhang-comparing-spanextra} \hfill + BERT & 87.38 & 87.70 & 88.15 & 87.93 & 81.52 & 81.36 & 81.44 & 86.27 & 86.00 & 86.84 & 86.42 \\
        \citet{generate-ijcai2021-0521} \hfill + BART & - & - & - & - & - & - & - & - & \bf87.8\textcolor{white}{0} & 86.8\textcolor{white}{0} & 87.3\textcolor{white}{0} \\
        \citet{zhang-2022-srlasdep} \hfill + BERT & \bf88.05 & 89.00 & \bf89.03 & \bf89.02 & 82.81 & \bf82.35 & \bf82.58 & \bf87.52 & 87.52 & \bf87.79 & \bf87.66 \\
        Ours \hfill + BERT & 87.54 & \bf89.03 & 88.53 & 88.78 & \bf83.22 & 81.81 & 82.51 & 86.97 & 87.26 & 87.05 & 87.15 \\
        \bottomrule
    \end{tabular}
    \caption{Results on CoNLL05 and CoNLL12 datasets. %
    We mark BIO-based models by $\dagger$ and span-based graph ones by $\ddagger$.
    \textcolor{black}{For \citet{strubell-2018-linguistically} and \citet{zhou-2020-parsing}, we list their syntax-agnostic results to compare fairly.}
    Moreover, we mark the results of \citet{strubell-2018-linguistically} by $^\ast$ to indicate that we report corrected evaluation results after re-testing their released models.
    } 
    \label{table:all results}
\end{small}
\end{table*}

Table \ref{table:speed} compares different models in terms of decoding speed. For fair comparison, we re-run all previous models on the same GPU environment (Nvidia GeForce 1080 Ti 11G). The results are averaged over 3 runs.
In terms of batch size during evaluation, our model and  \citet{strubell-2018-linguistically} use 5000 tokens (about 134 sentences), while \citet{he-2018-jointly} and \citet{Li-2019-dependency} use 40 sentences by default. 

We can see that our model improves the efficiency of previous span-based SRL models by large margin.
Compared with the span-based graph parsing approach \citep{he-2018-jointly,Li-2019-dependency},
our graph-based parser only has a $O(n^2)$ search space. 
As for the BIO-based model of \citet{strubell-2018-linguistically}, the encoder contains $12$ self-attention layers, and they adopts a pipeline framework by first predicting all predicates via sequence labeling and then recognizing arguments, leading to its low parsing speed. 
And when augmented with BERT, our methods can still parse about 250 sentences per second.

As discussed in Section \ref{related work},  \citet{zhang-2022-srlasdep} reduce the SRL to a tree parsing task and get good results.  
However, they have to build a dependency tree for each predicate, which greatly reduces the efficiency of their approach. Specifically, the speed of our model is respectively three and twice times as fast as theirs under the setting without and with BERT. 
 
\subsection{Comparison with previous results}
\paragraph{End-to-end.} 
Our work mainly focuses on the \emph{end-to-end} setting, i.e, requiring predicting predicates and arguments simultaneously.
So we first go into this scenario.
The first part of the Table \ref{table:all results} shows the comparison with previous works under the \emph{end-to-end} setting.

First, when compared with models without PLMs, our model surpasses previous approaches with the large gap, getting comparable results with recently released work \citep{zhang-2022-srlasdep}.
Then, most previous works usually use ELMo to improve the performance.
In order to make a fair comparison, we also report the results with ELMo.
\textcolor{black}{We can find that our model also reaches better results, with +0.25 $\textrm{F}_1$ on WSJ, +0.77 $\textrm{F}_1$ on Brown, and +0.44 $\textrm{F}_1$ on CoNLL12-test when using ELMo.}
And when augmented with the more powerful PLM BERT, the performance of our model can be further improved.
It shows that our method not only has high efficiency, but also performs better than previous works.

As discussed in Section \ref{related work}, please kindly notice that \citet{strubell-2018-linguistically} and \citet{zhou-2020-parsing} use 
 extra syntactic knowledge to boost SRL performance. 
We only list their syntax-agnostic  results here for fair comparison. 
It is worth noting that \citet{strubell-2018-linguistically} 
incidentally wrongly used the official script in the end-to-end setting, leading to much higher precision scores. 
We reported this issue to their github repository and they confirmed this mistake. 
In this work, we report their results by evaluating  their released models with the correct evaluation process.

\paragraph{Predicate-given.} 
Recent works \citep{jindal-2020-improved,zhang-comparing-spanextra, generate-ijcai2021-0521}
usually assume that predicates have been given, thus they only need to recognize the arguments and semantic roles.
To compare with these works, we also report the results under the \emph{predicate-given} setting.
In our work, following \citet{cai-2018-full}, during training procedure, the model is informed which word is the predicate using a predicate embedding. The embedding is added to the input vector.

Finally, from the second part of the Table \ref{table:all results}, we can see that our model reaches the best results on most test datasets when compared with models without PLMs.
When it comes to models with PLMs, we can see that the BIO-based \citet{shi-2019-simple} is a strong baseline.
Our model lags behind them slightly on WSJ, but is much higher than them on other datasets.
And even compared with recent seq-to-seq model \citep{generate-ijcai2021-0521}, which uses more powerful BART \citep{bart-2020}, our model still has strong competitiveness.

\textbf{Comparison with \citet{zhang-2022-srlasdep}.} 
As discussed in Section \ref{related work}, \citet{zhang-2022-srlasdep} propose a tree parsing approach to span-based SRL, which also appears in COLING-2022. 
We can see that performance of our model is slightly inferior to theirs, 
possibly due to more careful hyper-parameter tuning according to personal discussion between the two first authors. For example, Zhang confirms that fine-tuning BERT for 20 iterations leads to higher performance, while we only did 10 iterations.

\section{Conclusions}
\textcolor{black}{This paper proposes four new graph representation schemata for transforming raw span-based SRL structures to word-based graphs. Based on the schema, we cast the span-based SRL as a word-based graph parsing task and present a fast and accurate parser. Moreover, we propose a simple post-processing method based on constrained Viterbi to handle conflicts in the output graphs.  Experiments show that our parser 1) is much more efficient than previous parsers, and can parse over 600 sentences per second; 2) reaches consistently better performance than previous results on CoNLL05, CoNLL12 datasets. 
The in-depth comparison between four schemata shows that the boundary information counts a lot when recognizing arguments.  
In addition, distinguishing single-word arguments from multi-words arguments can also improve the final performance.
These clear findings may help researchers think about SRL from a new perspective in the future. 
}

\section{Acknowledgments}
We thank our anonymous reviewers for their valuable suggestions.
This work was supported by National
Natural Science Foundation of China (No. 62176173 and 62076174) and a project funded by the Priority Academic Program Development of Jiangsu Higher Education Institutions.

\bibliography{anthology}

\begin{thebibliography}{33}
\expandafter\ifx\csname natexlab\endcsname\relax\def\natexlab#1{#1}\fi

\bibitem[{Blloshmi et~al.(2021)Blloshmi, Conia, Tripodi, and
  Navigli}]{generate-ijcai2021-0521}
Rexhina Blloshmi, Simone Conia, Rocco Tripodi, and Roberto Navigli. 2021.
\newblock Generating senses and roles: An end-to-end model for dependency- and
  span-based semantic role labeling.
\newblock In \emph{Proceedings of IJCAI}, pages 3786--3793.

\bibitem[{Cai et~al.(2018)Cai, He, Li, and Zhao}]{cai-2018-full}
Jiaxun Cai, Shexia He, Zuchao Li, and Hai Zhao. 2018.
\newblock A full end-to-end semantic role labeler, syntactic-agnostic over
  syntactic-aware?
\newblock In \emph{Proceedings of ACL}, pages 2753--2765.

\bibitem[{Devlin et~al.(2019)Devlin, Chang, Lee, and
  Toutanova}]{devlin-2018-bert}
Jacob Devlin, Ming-Wei Chang, Kenton Lee, and Kristina Toutanova. 2019.
\newblock Bert: Pre-training of deep bidirectional transformers for language
  understanding.
\newblock In \emph{Proceedings of NAACL-HLT}, pages 4171--4186.

\bibitem[{Dozat and Manning(2018)}]{dozat-2018-simpler}
Timothy Dozat and Christopher~D. Manning. 2018.
\newblock Simpler but more accurate semantic dependency parsing.
\newblock In \emph{Proceedings of ACL}, pages 484--490.

\bibitem[{Haji{\v{c}} et~al.(2012)Haji{\v{c}}, Haji{\v{c}}ov{\'a},
  Panevov{\'a}, Sgall, Bojar, Cinkov{\'a}, Fu{\v{c}}{\'\i}kov{\'a},
  Mikulov{\'a}, Pajas, Popelka, Semeck{\'y}, {\v{S}}indlerov{\'a},
  {\v{S}}t{\v{e}}p{\'a}nek, Toman, Ure{\v{s}}ov{\'a}, and
  {\v{Z}}abokrtsk{\'y}}]{hajic-etal-2012-psd}
Jan Haji{\v{c}}, Eva Haji{\v{c}}ov{\'a}, Jarmila Panevov{\'a}, Petr Sgall,
  Ond{\v{r}}ej Bojar, Silvie Cinkov{\'a}, Eva Fu{\v{c}}{\'\i}kov{\'a}, Marie
  Mikulov{\'a}, Petr Pajas, Jan Popelka, Ji{\v{r}}{\'\i} Semeck{\'y}, Jana
  {\v{S}}indlerov{\'a}, Jan {\v{S}}t{\v{e}}p{\'a}nek, Josef Toman, Zde{\v{n}}ka
  Ure{\v{s}}ov{\'a}, and Zden{\v{e}}k {\v{Z}}abokrtsk{\'y}. 2012.
\newblock Announcing {P}rague {C}zech-{E}nglish {D}ependency {T}reebank 2.0.
\newblock In \emph{Proceedings of LREC}, pages 3153--3160.

\bibitem[{He et~al.(2018)He, Lee, Levy, and Zettlemoyer}]{he-2018-jointly}
Luheng He, Kenton Lee, Omer Levy, and Luke Zettlemoyer. 2018.
\newblock Jointly predicting predicates and arguments in neural semantic role
  labeling.
\newblock In \emph{Proceedings of ACL}, pages 364--369.

\bibitem[{He et~al.(2017)He, Lee, Lewis, and Zettlemoyer}]{he-etal-2017-deep}
Luheng He, Kenton Lee, Mike Lewis, and Luke Zettlemoyer. 2017.
\newblock Deep semantic role labeling: What works and what{'}s next.
\newblock In \emph{Proceedings of ACL}, pages 473--483.

\bibitem[{Ivanova et~al.(2012)Ivanova, Oepen, {\O}vrelid, and
  Flickinger}]{ivanova-etal-2012-dm}
Angelina Ivanova, Stephan Oepen, Lilja {\O}vrelid, and Dan Flickinger. 2012.
\newblock Who did what to whom? {A} contrastive study of syntacto-semantic
  dependencies.
\newblock In \emph{Proceedings of LAW}, pages 2--11.

\bibitem[{Jindal et~al.(2020)Jindal, Aharonov, Brahma, Zhu, and
  Li}]{jindal-2020-improved}
Ishan Jindal, Ranit Aharonov, Siddhartha Brahma, Huaiyu Zhu, and Yunyao Li.
  2020.
\newblock Improved semantic role labeling using parameterized neighborhood
  memory adaptation.
\newblock \emph{arXiv preprint arXiv:2011.14459}.

\bibitem[{Lample et~al.(2016)Lample, Ballesteros, Subramanian, Kawakami, and
  Dyer}]{lample-2016-neural}
Guillaume Lample, Miguel Ballesteros, Sandeep Subramanian, Kazuya Kawakami, and
  Chris Dyer. 2016.
\newblock Neural architectures for named entity recognition.
\newblock In \emph{Proceedings of NAACL-HLT}, pages 260--270.

\bibitem[{Lewis et~al.(2020)Lewis, Liu, Goyal, Ghazvininejad, Mohamed, Levy,
  Stoyanov, and Zettlemoyer}]{bart-2020}
Mike Lewis, Yinhan Liu, Naman Goyal, Marjan Ghazvininejad, Abdelrahman Mohamed,
  Omer Levy, Veselin Stoyanov, and Luke Zettlemoyer. 2020.
\newblock {BART}: Denoising sequence-to-sequence pre-training for natural
  language generation, translation, and comprehension.
\newblock In \emph{Proceedings of ACL}, pages 7871--7880.

\bibitem[{Li et~al.(2019)Li, He, Zhao, Zhang, Zhang, Zhou, and
  Zhou}]{Li-2019-dependency}
Zuchao Li, Shexia He, Hai Zhao, Yiqing Zhang, Zhuosheng Zhang, Xi~Zhou, and
  Xiang Zhou. 2019.
\newblock Dependency or span, end-to-end uniform semantic role labeling.
\newblock In \emph{Proceedings of AAAI}, pages 6730--6737.

\bibitem[{Li et~al.(2020)Li, Zhao, Wang, and Parnow}]{li-2020-high}
Zuchao Li, Hai Zhao, Rui Wang, and Kevin Parnow. 2020.
\newblock High-order semantic role labeling.
\newblock In \emph{Findings of EMNLP}, pages 1134--1151.

\bibitem[{Liu and Gildea(2010)}]{liu-2010-semantic}
Ding Liu and Daniel Gildea. 2010.
\newblock Semantic role features for machine translation.
\newblock In \emph{Proceedings of COLING}, pages 716--724.

\bibitem[{Marcheggiani et~al.(2018)Marcheggiani, Bastings, and
  Titov}]{marcheggiani-2018-exploiting}
Diego Marcheggiani, Jasmijn Bastings, and Ivan Titov. 2018.
\newblock Exploiting semantics in neural machine translation with graph
  convolutional networks.
\newblock In \emph{Proceedings of NAACL-HLT}, pages 486--492.

\bibitem[{Miyao and Tsujii(2004)}]{miyao-2004-pas}
Yusuke Miyao and Jun{'}ichi Tsujii. 2004.
\newblock Deep linguistic analysis for the accurate identification of
  predicate-argument relations.
\newblock In \emph{Proceedings of COLING}.

\bibitem[{Oepen et~al.(2015)Oepen, Kuhlmann, Miyao, Zeman, Cinkov{\'a},
  Flickinger, Hajic, and Uresova}]{oepen-2015-semeval}
Stephan Oepen, Marco Kuhlmann, Yusuke Miyao, Daniel Zeman, Silvie Cinkov{\'a},
  Dan Flickinger, Jan Hajic, and Zdenka Uresova. 2015.
\newblock Semeval 2015 task 18: Broad-coverage semantic dependency parsing.
\newblock In \emph{Proceedings of SemEval}, pages 915--926.

\bibitem[{Oepen et~al.(2014)Oepen, Kuhlmann, Miyao, Zeman, Flickinger,
  Haji{\v{c}}, Ivanova, and Zhang}]{oepen-2014-semeval}
Stephan Oepen, Marco Kuhlmann, Yusuke Miyao, Daniel Zeman, Dan Flickinger, Jan
  Haji{\v{c}}, Angelina Ivanova, and Yi~Zhang. 2014.
\newblock {S}em{E}val 2014 task 8: Broad-coverage semantic dependency parsing.
\newblock In \emph{Proceedings of SemEval}, pages 63--72.

\bibitem[{Palmer et~al.(2005)Palmer, Gildea, and
  Kingsbury}]{palmer-etal-2005-proposition}
Martha Palmer, Daniel Gildea, and Paul Kingsbury. 2005.
\newblock The proposition bank: An annotated corpus of semantic roles.
\newblock \emph{Computational linguistics}, 31(1):71--106.

\bibitem[{Pennington et~al.(2014)Pennington, Socher, and
  Manning}]{pennington-etal-2014-glove}
Jeffrey Pennington, Richard Socher, and Christopher Manning. 2014.
\newblock {G}lo{V}e: Global vectors for word representation.
\newblock In \emph{Proceedings of EMNLP}, pages 1532--1543.

\bibitem[{Peters et~al.(2018)Peters, Neumann, Iyyer, Gardner, Clark, Lee, and
  Zettlemoyer}]{peters-etal-2018-elmo}
Matthew~E. Peters, Mark Neumann, Mohit Iyyer, Matt Gardner, Christopher Clark,
  Kenton Lee, and Luke Zettlemoyer. 2018.
\newblock Deep contextualized word representations.
\newblock In \emph{Proceedings of NAACL-HLT}, pages 2227--2237.

\bibitem[{Pradhan et~al.(2012)Pradhan, Moschitti, Xue, Uryupina, and
  Zhang}]{pradhan-etal-2012-conll}
Sameer Pradhan, Alessandro Moschitti, Nianwen Xue, Olga Uryupina, and Yuchen
  Zhang. 2012.
\newblock {C}o{NLL}-2012 shared task: Modeling multilingual unrestricted
  coreference in {O}nto{N}otes.
\newblock In \emph{Proceedings of EMNLP-{C}o{NLL}}, pages 1--40.

\bibitem[{Shi and Lin(2019)}]{shi-2019-simple}
Peng Shi and Jimmy Lin. 2019.
\newblock Simple bert models for relation extraction and semantic role
  labeling.
\newblock \emph{arXiv preprint arXiv:1904.05255}.

\bibitem[{Strubell et~al.(2018)Strubell, Verga, Andor, Weiss, and
  McCallum}]{strubell-2018-linguistically}
Emma Strubell, Patrick Verga, Daniel Andor, David Weiss, and Andrew McCallum.
  2018.
\newblock Linguistically-informed self-attention for semantic role labeling.
\newblock In \emph{Proceedings of EMNLP}, pages 5027--5038.

\bibitem[{Tan et~al.(2018)Tan, Wang, Xie, Chen, and Shi}]{tan-2018-deep}
Zhixing Tan, Mingxuan Wang, Jun Xie, Yidong Chen, and Xiaodong Shi. 2018.
\newblock Deep semantic role labeling with self-attention.
\newblock In \emph{Proceedings of AAAI}.

\bibitem[{Wang et~al.(2015)Wang, Bansal, Gimpel, and
  McAllester}]{wang-2015-machine}
Hai Wang, Mohit Bansal, Kevin Gimpel, and David McAllester. 2015.
\newblock Machine comprehension with syntax, frames, and semantics.
\newblock In \emph{Proceedings of ACL-IJCNLP}, pages 700--706.

\bibitem[{Wang et~al.(2019)Wang, Huang, and Tu}]{tukewei-2019-2osdp}
Xinyu Wang, Jingxian Huang, and Kewei Tu. 2019.
\newblock Second-order semantic dependency parsing with end-to-end neural
  networks.
\newblock In \emph{Proceedings of ACL}, pages 4609--4618.

\bibitem[{Xia et~al.(2019)Xia, Li, Zhang, Zhang, Fu, Wang, and
  Si}]{xia-2019-syntax}
Qingrong Xia, Zhenghua Li, Min Zhang, Meishan Zhang, Guohong Fu, Rui Wang, and
  Luo Si. 2019.
\newblock Syntax-aware neural semantic role labeling.
\newblock In \emph{Proceedings of AAAI}, pages 7305--7313.

\bibitem[{Zhang et~al.(2020)Zhang, Li, and Zhang}]{zhang-2020-efficient}
Yu~Zhang, Zhenghua Li, and Min Zhang. 2020.
\newblock Efficient second-order {T}ree{CRF} for neural dependency parsing.
\newblock In \emph{Proceedings of ACL}, pages 3295--3305.

\bibitem[{Zhang et~al.(2022)Zhang, Xia, Zhou, Jiang, Fu, and
  Zhang}]{zhang-2022-srlasdep}
Yu~Zhang, Qingrong Xia, Shilin Zhou, Yong Jiang, Guohong Fu, and Min Zhang.
  2022.
\newblock Semantic role labeling as dependency parsing: Exploring latent tree
  structures inside arguments.
\newblock In \emph{Proceedings of COLING}.

\bibitem[{Zhang et~al.(2021)Zhang, Strubell, and
  Hovy}]{zhang-comparing-spanextra}
Zhisong Zhang, Emma Strubell, and Eduard Hovy. 2021.
\newblock Comparing span extraction methods for semantic role labeling.
\newblock In \emph{Proceedings of SPNLP}, pages 67--77.

\bibitem[{Zhou and Xu(2015)}]{zhou-2015-end}
Jie Zhou and Wei Xu. 2015.
\newblock End-to-end learning of semantic role labeling using recurrent neural
  networks.
\newblock In \emph{Proceedings of ACL-IJCNLP}, pages 1127--1137.

\bibitem[{Zhou et~al.(2020)Zhou, Li, and Zhao}]{zhou-2020-parsing}
Junru Zhou, Zuchao Li, and Hai Zhao. 2020.
\newblock Parsing all: Syntax and semantics, dependencies and spans.
\newblock In \emph{Findings of EMNLP}, pages 4438--4449.

\end{thebibliography}
\bibliographystyle{acl_natbib}

\appendix

\vspace{+2ex}
\begin{center}
\Large \textbf{Appendices} 
\end{center}
\vspace{+1ex}

\section{More examples of other schemata} \label{app:other_matrix}
\begin{figure}[htb]
    \centering

    \includegraphics[width=0.4\columnwidth]{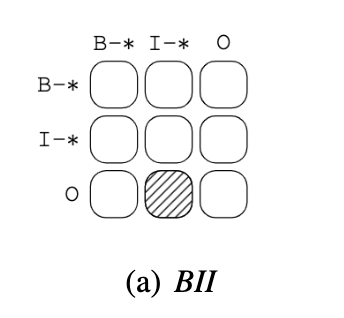}
    \includegraphics[width=0.4\columnwidth]{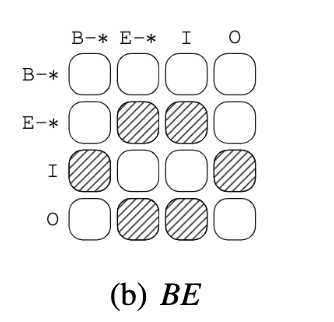}
    \includegraphics[width=0.5\columnwidth]{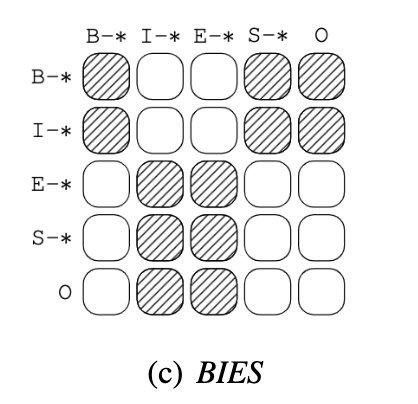}
    
    \caption{Transition matrices of \emph{BII}, \emph{BE}, and \emph{BIES}.}
    \label{fig:other matrix}
\end{figure}

\begin{figure}[htb]
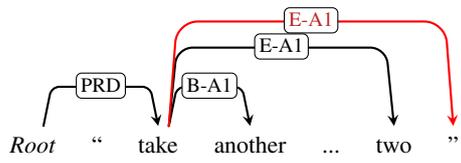
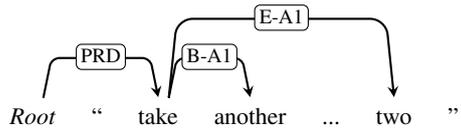

    \centering
    \subfigure[w/o Constrained Viterbi]{
    \scalebox{1}{
    \begin{minipage}[t]{1\columnwidth}
            \centering
            \begin{dependency}[arc angle=80]
                \begin{deptext}[row sep=0.2cm, column sep=.3cm, font=\small]
                            \emph{Root} \& `` \& take \& another \& ... \& two \& '' \\
                \end{deptext}
                \depedge[edge style={black,thick}, edge height=3ex]{1}{3}{PRD}
                \depedge[edge style={black,thick}, edge height=3ex]{3}{4}{B-A1}
                \depedge[edge style={black,thick}, edge height=6ex]{3}{6}{E-A1}
                \depedge[edge style={red,thick}, edge height=8ex]{3}{7}{\red{E-A1}}
            \end{dependency}
        \end{minipage}%
        \label{fig:cft-BE-false}
    }
    }
    \subfigure[w/ Constrained Viterbi]{
    \scalebox{1}{
    \begin{minipage}[t]{1\columnwidth}
            \centering
            \begin{dependency}[arc angle=80]
                \begin{deptext}[row sep=0.2cm, column sep=.3cm, font=\small]
                            \emph{Root} \& `` \& take \& another \& ... \& two \& '' \\
                \end{deptext}
                \depedge[edge style={black,thick}, edge height=3ex]{1}{3}{PRD}
                \depedge[edge style={black,thick}, edge height=3ex]{3}{4}{B-A1}
                \depedge[edge style={black,thick}, edge height=6ex]{3}{6}{E-A1}
            \end{dependency}
        \end{minipage}%
        }
        \label{fig:cft-BE-true}
    }
    \caption{\emph{BE} schema.}
    \label{fig:cft-BE}
\end{figure}

\begin{figure}[htb]
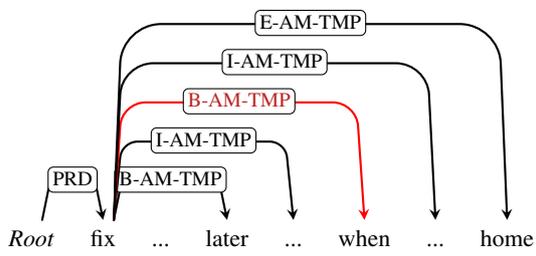
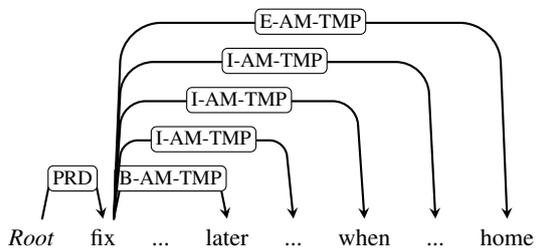

    \centering
    \subfigure[w/o Constrained Viterbi]{
    \scalebox{1}{
    \begin{minipage}[t]{1\columnwidth}
            \centering
            \begin{dependency}[arc angle=80]
                \begin{deptext}[row sep=0.2cm, column sep=.3cm, font=\small]
                            \emph{Root} \& fix \& ... \& later \& ... \& when \& ... \& home \\
                \end{deptext}
                \depedge[edge style={black,thick}]{1}{2}{PRD}
                \depedge[edge style={black,thick},edge height=3ex]{2}{4}{B-AM-TMP}
                \depedge[edge style={black,thick},edge height=6ex]{2}{5}{I-AM-TMP}
                \depedge[edge style={red,thick},edge height=9ex]{2}{6}{\red{B-AM-TMP}}
                \depedge[edge style={black,thick},edge height=12ex]{2}{7}{I-AM-TMP}
                \depedge[edge style={black,thick},edge height=15ex]{2}{8}{E-AM-TMP}
            \end{dependency}
        \end{minipage}%
        \label{fig:cft-BIES-false}
    }
    }
    \subfigure[w/ Constrained Viterbi]{
    \scalebox{1}{
    \begin{minipage}[t]{1\columnwidth}
            \centering
            \begin{dependency}[arc angle=80]
                \begin{deptext}[row sep=0.2cm, column sep=.3cm, font=\small]
                            \emph{Root} \& fix \& ... \& later \& ... \& when \& ... \& home \\
                \end{deptext}
                \depedge[edge style={black,thick}]{1}{2}{PRD}
                \depedge[edge style={black,thick},edge height=3ex]{2}{4}{B-AM-TMP}
                \depedge[edge style={black,thick},edge height=6ex]{2}{5}{I-AM-TMP}
                \depedge[edge style={black,thick},edge height=9ex]{2}{6}{I-AM-TMP}
                \depedge[edge style={black,thick},edge height=12ex]{2}{7}{I-AM-TMP}
                \depedge[edge style={black,thick},edge height=15ex]{2}{8}{E-AM-TMP}
            \end{dependency}
        \end{minipage}%
        }
        \label{fig:cft-BIES-true}
    }
    \caption{\emph{BIES} schema.}
    \label{fig:cft-BIES}
\end{figure}

In order to save space, we only show the transition matrix of \emph{BES} in the main body.
Here, we present the transition matrix of others in Figure \ref{fig:other matrix}.
In addition, we provide more examples to improve the comprehensibility of our schemata and the constrained viterbi.
Figure \ref{fig:cft-BII}, \ref{fig:cft-BE}, and \ref{fig:cft-BIES} show the outputs of models using different schemata respectively. 
For example, in Figure \ref{fig:cft-BII}, there is missing an edge from ``pilling'' to ``falling'' in the raw output.
After the viterbi procedure, an \texttt{I-AM-ADV} edge will be added since we disallow the transition from \texttt{O} to \texttt{I-$\ast$}. Thus we can get the legal SRL structure.

\begin{figure}[htb]
    \centering
    \subfigure[w/o Constrained Viterbi]{
    \scalebox{1}{
    \begin{minipage}[t]{1\columnwidth}
            \centering
            \begin{dependency}[arc angle=80]
                \begin{deptext}[row sep=0.2cm, column sep=.3cm, font=\small]
                            \emph{Root} \& pilling \& ... \& while \& falling \& sharply \\
                \end{deptext}
                \depedge[edge style={black,thick}]{1}{2}{PRD}
                \depedge[edge style={black,thick}, edge height=3ex]{2}{4}{B-AM-ADV}
                \depedge[edge style={red,dashed}, edge height=6ex]{2}{5}{\red{I-AM-ADV}}
                \depedge[edge style={black,thick}, edge height=8ex]{2}{6}{I-AM-ADV}
            \end{dependency}
        \end{minipage}%
        \label{fig:cft-BII-false}
    }
    }
    \subfigure[w/ Constrained Viterbi]{
    \scalebox{1}{
    \begin{minipage}[t]{1\columnwidth}
            \centering
            \begin{dependency}[arc angle=80]
                \begin{deptext}[row sep=0.2cm, column sep=.3cm, font=\small]
                            \emph{Root} \& pilling \& ... \& while \& falling \& sharply \\
                \end{deptext}
                \depedge[edge style={black,thick}]{1}{2}{PRD}
                \depedge[edge style={black,thick}, edge height=3ex]{2}{4}{B-AM-ADV}
                \depedge[edge style={black,thick}, edge height=6ex]{2}{5}{I-AM-ADV}
                \depedge[edge style={black,thick}, edge height=8ex]{2}{6}{I-AM-ADV}
            \end{dependency}
        \end{minipage}%
        }
        \label{fig:cft-BII-true}
    }
    \caption{\emph{BII} schema.}
    \label{fig:cft-BII}
\end{figure}

\end{CJK}
\end{document}